\pdfoutput=1

\documentclass[11pt]{article}

\usepackage[]{acl}

\usepackage{times}
\usepackage{latexsym}
\usepackage{graphicx}

\usepackage[T1]{fontenc}

\usepackage[utf8]{inputenc}

\usepackage{microtype}

\usepackage{times}
\usepackage{latexsym}
\usepackage{tcolorbox}
\usepackage{multirow}
\usepackage{tikz}
\usepackage{listings}
\usepackage{capt-of}
\usepackage{graphicx}  
\usepackage{pgfplots}
\usepackage{overpic}
\pgfplotsset{compat=1.12}
\usepackage{amsmath}
\usepackage{multicol}
\usepackage{booktabs}
\usepackage{colortbl,array,xcolor}
\usepackage{enumitem}
\usepackage{xspace}
\usepackage{caption}
\usepackage{subcaption}
\usepackage{graphicx}
\usepackage{amsfonts}
\usepackage{booktabs}
\usepackage{xcolor}
\usepackage{soul}
\usepackage{graphicx}
\usepackage{xspace}
\usepackage{multirow} 
\usepackage{amsmath}

\newcommand{\ours}{\textsc{PopQA}}
\usetikzlibrary{intersections}

\newcommand{\subscript}[2]{$#1 _ #2$}

\newcommand{\hlc}[2][yellow]{{%
    \colorlet{foo}{#1}%
    \sethlcolor{foo}\hl{#2}}%
}

\definecolor{Gray}{gray}{0.85}
\definecolor{LightCyan}{rgb}{0.88,1,1}

\newcolumntype{a}{>{\columncolor{Gray}}c}
\newcolumntype{b}{>{\columncolor{LightCyan}}c}

\usetikzlibrary{shapes.geometric}

\newcommand{\draftonly}[1]{#1}
\newcommand{\draftcomment}[3]{\draftonly{{\textcolor{#3}{[\textbf{#1--\textsc{#2}}]}}}}

\newcommand{\akari}[1]{\draftcomment{\small #1}{akari}{purple}}

\usepackage{pifont}
\newcommand{\xmark}{\ding{55}}
\usepackage{algorithm}
\usepackage{algpseudocode}%

\title{\vspace*{-0.5in}
{{\small \hfill ACL 2023}\\
\vspace*{.25in}}
When Not to Trust Language Models: Investigating Effectiveness of Parametric and Non-Parametric Memories}

\author{\parbox{0.9\linewidth}{
\centering{Alex Mallen$^{*\diamondsuit}$ ~~~ Akari Asai$^{*\diamondsuit}$ ~~~ Victor Zhong$^{\diamondsuit}$ ~~~ Rajarshi Das$^{\diamondsuit}$ \\
Daniel Khashabi$^\spadesuit$ ~~~ Hannaneh Hajishirzi$^{\diamondsuit\heartsuit}$ 
} \\
{\rm $^\diamondsuit$University of Washington~~$^\spadesuit$Johns Hopkins University\\$^\heartsuit$Allen Institute for AI} \\
\texttt{\{atmallen,akari,vzhong,rajarshi,hannaneh\}@cs.washington.edu} \\
\texttt{danielk@jhu.edu} \\
}
}

\begin{document}
\maketitle

\begin{abstract}
Despite their impressive performance on diverse tasks, large language models (LMs) still struggle with tasks requiring rich world knowledge, implying the difficulty of encoding a wealth of world knowledge in their parameters. 
This paper aims to understand LMs' strengths and limitations in memorizing factual knowledge, by conducting large-scale knowledge probing experiments on two open-domain entity-centric QA datasets: \ours, our new dataset with 14k questions about long-tail entities, and EntityQuestions, a widely used open-domain QA dataset.
We find that LMs struggle with less popular factual knowledge, and that retrieval augmentation helps significantly in these cases. Scaling, on the other hand, mainly improves memorization of popular knowledge, and fails to appreciably improve memorization of factual knowledge in the long tail.
Based on those findings, we devise a new method for retrieval augmentation that improves performance and reduces inference costs by only retrieving non-parametric memories when necessary.\footnote{Our code and data are available at \url{https://github.com/AlexTMallen/adaptive-retrieval}. }

\end{abstract}

\section{Introduction}
Large language models (LMs; ~\citealt{brown2020language,raffel2020exploring}) have been shown to be competitive on diverse NLP tasks, including knowledge-intensive tasks that require fine-grained memorization of factual knowledge~\cite{chowdhery2022palm,yu2022generate}.  
Meanwhile, LMs have also been shown to have limited memorization for less frequent entities~\cite{llm_longtail}, 
are prone to hallucinations~\cite{shuster2021retrieval}, and suffer from temporal degradation~\cite{kasai2022realtime,jang2022temporalwiki}.
Incorporating \emph{non-parametric knowledge} (i.e., retrieved text chunks) largely helps address those issues stemming from reliance on LMs' \emph{parametric knowledge}---knowledge stored in their parameters~\cite{izacard2022few}---but it is unclear whether it is strictly superior or complementary to parametric knowledge. 
Understanding when we should \emph{not} trust LMs' outputs is also crucial to safely deploying them in real-world applications~\cite{kadavath2022language}.

\begin{figure}[t!]
\includegraphics[width=7.6cm, trim=0cm 0.6cm 0cm 0.2cm]{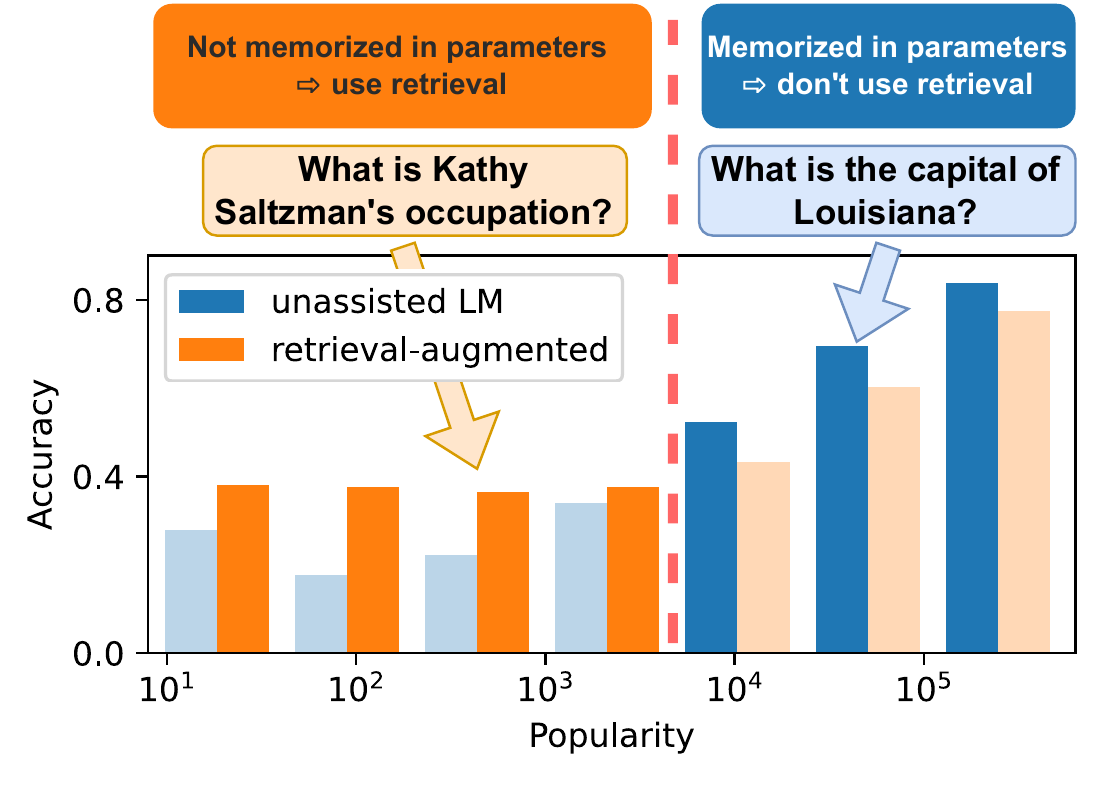}\caption{
Relationship between subject entity popularity in a question and GPT-3 performance in open-domain QA, with and without retrieved passages. Adaptive Retrieval only retrieves when necessary (orange bars) based on the heuristically-decided threshold (red line). 
} \label{fig:teaser}
\end{figure}

This work conducts a large-scale knowledge probing of LMs on factual knowledge memorization, to understand when we should and should \emph{not} rely on LMs' parametric knowledge, and how scaling and non-parametric memories (e.g., retrieval-augmented LMs) can help.
In particular, we aim to address the following research questions: 
\vspace{-0.2cm}
\begin{enumerate}[label=(\subscript{RQ}{{\arabic*}}),leftmargin=35pt]
\itemsep-0.4em 
    \item \label{Q1}  
    How much factual knowledge is memorized by LMs and what factors affect the memorization? (Section~\ref{sec:analysis_parametric})
    \item \label{Q2}  To what extent can non-parametric memories alleviate the shortcomings of parametric memories of LMs?
    (Section~\ref{sec:analysis_non_parametric})
    \item \label{Q3}  Can we build a system to adaptively combine non-parametric and parametric memories? (Section~\ref{sec:adaptive_retrieval})
\end{enumerate}
\vspace{-0.2cm}

We hypothesize that factual knowledge frequently discussed on the web is easily memorized by LMs, while the knowledge that is less discussed may not be well captured and thus they require retrieving external non-parametric memories.
{
We evaluate ten large LMs of three families (i.e., GPT-Neo, OPT, and GPT-3) with varying scales on the open-domain question answering (QA) task in a zero- or few-shot prompting manner.} 
We construct a new dataset, \ours, consisting of 14k questions to cover factual information in the long tail that might have been missed in popular QA datasets~\cite{kwiatkowski-etal-2019-natural}. 
{We use Wikipedia page views as a measure of popularity and convert knowledge triples from Wikidata, with diverse levels of popularity, into natural language questions, anchored to the original entities and relationship types.}
{We also use EntityQuestions~\cite{sciavolino2021simple}, an open-domain QA dataset with a long-tail distribution. 
}

On both datasets, LMs' memorization {\ref{Q1}} is often limited to the popular factual knowledge and even GPT-3 \texttt{davinci-003} fails to answer the majority of the long-tail questions. 
Moreover, on such questions, scaling up models does \emph{not} significantly improve the performance (e.g., for the 4,000 least popular questions in \ours, GPT-j 6B has 16\% accuracy and GPT-3 \texttt{davinci-003} has 19\% accuracy). 
{This also suggests that we can predict if LMs memorize certain knowledge based on the information presented in the input question only. }

We next investigate whether a semi-parametric approach that augments LMs with retrieved evidence can mitigate the low performance on questions about less popular entities {\ref{Q2}}. 
Non-parametric memories largely improve performance on long-tail distributions across models. 
Specifically, we found that retrieval-augmented LMs are particularly competitive when subject entities are not popular: {a neural dense retriever}~\cite{izacard2021towards}-augmented GPT-neo 2.7B outperforms GPT-3 \texttt{davinci-003} on the 4,000 least popular questions.
{Surprisingly, we also find that retrieval augmentation can hurt the performance of large LMs on questions about popular entities as the retrieved context can be misleading.} 

As a result, we devise a simple-yet-effective retrieval-augmented LM method, Adaptive Retrieval, which adaptively combines parametric and non-parametric memories based on {popularity}
{\ref{Q3}}.
This method further improves performance on \ours\ by up to {10\%, while significantly reducing the inference costs, especially with larger LMs (e.g., reducing GPT-3 API costs by half),  
{
indicating the potential for future research in more efficient and powerful retrieval-augmented LMs.
}

\section{Related Work}

\paragraph{Parametric and non-parametric knowledge.}
{\citet{petroni2019language} demonstrate that large pre-trained LMs such as BERT~\cite{devlin2018bert} memorize the significant amount of world knowledge in their parameters ({\it parametric knowledge}), and \citet{roberts2020much} show that fine-tuned T5 without any reference documents (closed-book QA) can achieve competitive performance on open-domain QA. 
More recent and powerful LMs~\cite{brown2020language,chowdhery2022palm} further improve performance on diverse knowledge-intensive tasks, leveraging their strong parametric memories~\cite{llm_longtail,yu2022generate}.}
However, relying solely on their parameters to encode a wealth of world knowledge requires a prohibitively large number of parameters and the knowledge can become obsolete quickly~\cite{kasai2022realtime,jang2022temporalwiki}. 
Recent work shows that augmenting LMs with non-parametric memories (i.e., retrieved text chunks) enables much smaller models to match the performance of larger models~\cite{izacard2022few,khandelwal2019generalization,min2022nonparametric}{, although \citet{chen2022rich} and \citet{longpre-etal-2021-entity} show that even those models can ignore non-parametric knowledge and rely on parametric knowledge.
}

\paragraph{Understanding memorization. }
Several prior work establishes a positive relationship between string frequency in pre-training corpora and memorization~\cite{Carlini2022QuantifyingMA,Razeghi2022ImpactOP}.
Concurrent to our work, ~\citet{llm_longtail} show that the co-occurrence of the question and answer entities in pretraining corpora has a positive correlation with models' QA accuracy on popular open-domain QA benchmarks such as Natural Questions~\cite{kwiatkowski-etal-2019-natural}.
This work, instead, attempts to predict memorization using the variables available in the input question only and uses popularity to obtain a proxy for how frequently an entity is likely to be discussed on the web.
Importantly, by constructing a new dataset, we can conduct fine-grained controlled experiments across a wide range of popularities, allowing the investigation of hypotheses that might have been missed in prior analysis using existing open QA datasets.
We further analyze the effectiveness and limitations of retrieval-augmented LMs and {introduce Adaptive Retrieval. 
}
Prior work investigates the effectiveness of deciding when to use non-parametric memories at the token level in $k$NN LM~\cite{he-etal-2021-efficient,drozdov-etal-2022-cant}. 
This work is the first work to study the effectiveness of deciding whether to retrieve for each query and show their effectiveness in retrieval-augmented LM prompting.

\begin{figure}[t!]
    \centering
\includegraphics[width=0.95\linewidth,keepaspectratio]{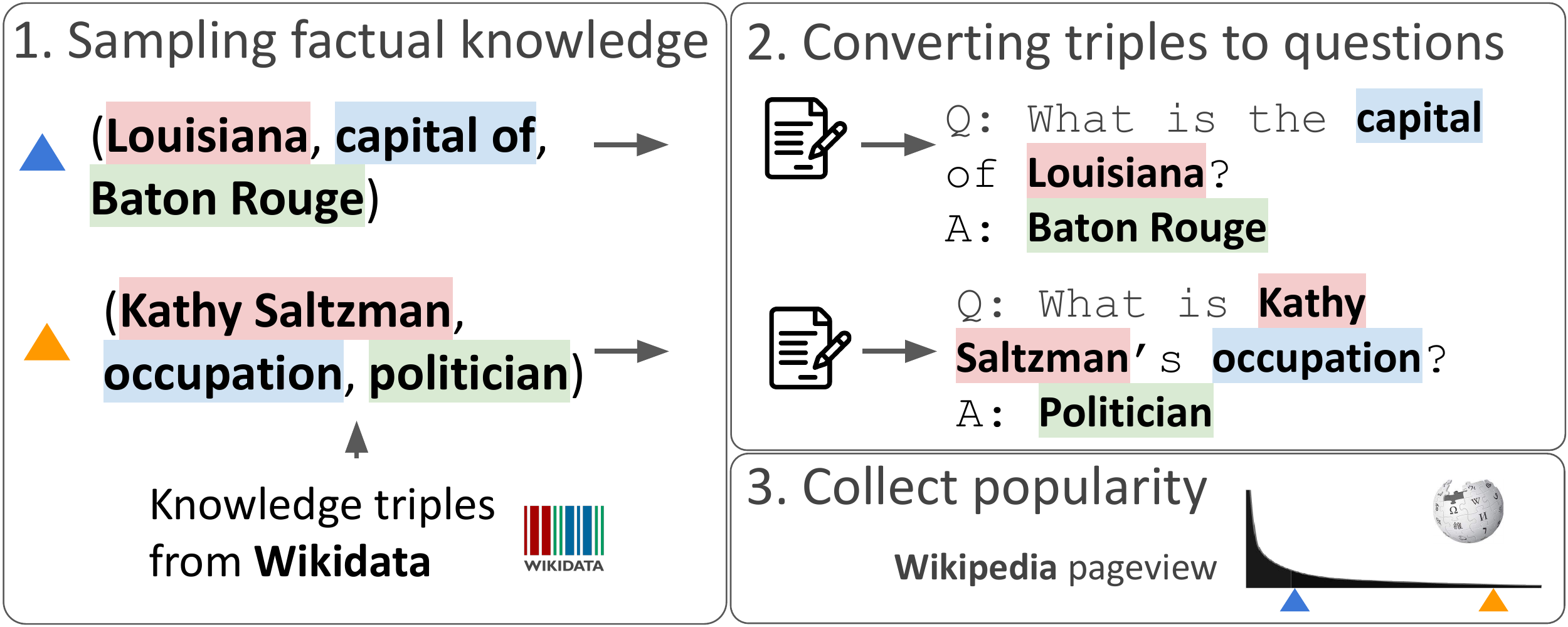}
    \caption{\ours~is created by sampling knowledge triples from Wikidata and converting them to natural language questions, followed by popularity calculation. 
    }
    \label{fig:background}
\end{figure}
\section{Evaluation Setup}
\label{sec:dataset}

{
We evaluate LMs' ability to memorize factual knowledge through closed-book QA tasks with few-shot samples. 
We evaluate LMs on our new dataset, \ours~(Figure~\ref{fig:background}), and EntityQuestions, both of which have long-tail distributions (Figure~\ref{fig:pop_distr}).
}

\begin{figure}[ht!]
    \centering
    \includegraphics[width=0.9\linewidth]{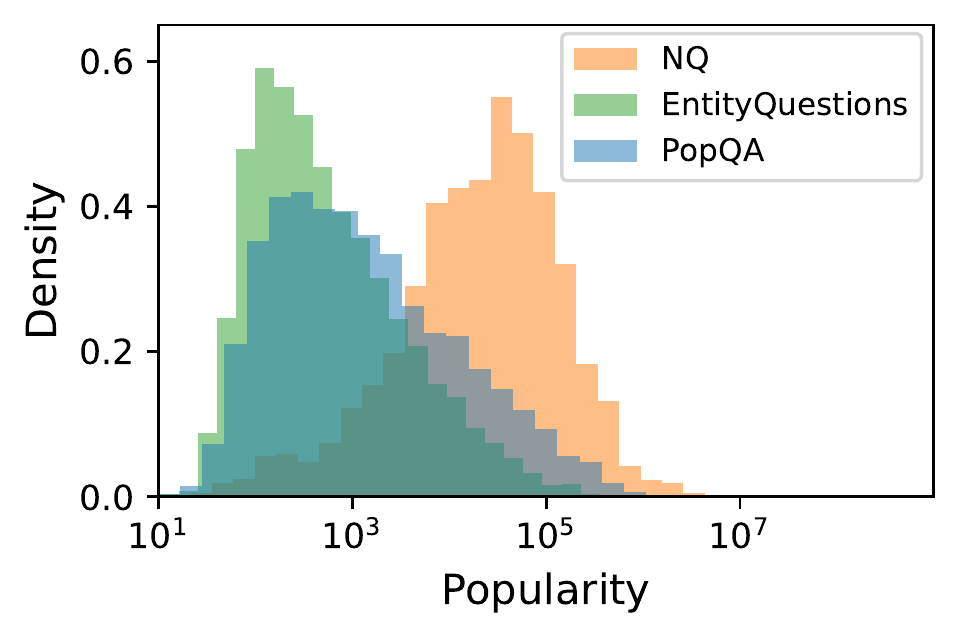}
    \caption{Distribution of subject entity popularity for EntityQuestions, \ours,\ and for NQ-open for reference. Details on NQ entities can be found in Appendix~\ref{app_sec:dataset}.
    }
    \label{fig:pop_distr}
\end{figure}

\subsection{Focus and Task}
\noindent {\bf Focus: factual knowledge. }
Among diverse types of world knowledge, this work focuses on factual knowledge~\cite{adams2015bloom} of entities---knowledge about specific details of the target entities. 
We define factual knowledge as a triplet of (\hlc[pink]{subject}, \hlc[cyan!20]{relationship}, \hlc[green!20]{object}) as in Figure~\ref{fig:background} left.

\vspace{.1cm}
\noindent {\bf Task format: open-domain QA. } 
We formulate the task as open-domain QA~\cite{roberts2020much}: given a question, a model predicts an answer without any pre-given ground-truth paragraph.\footnote{Some work conducts knowledge probing of encoder-only models by filling out \texttt{[MASK]} tokens~\cite{petroni2019language}. 
We use decoder-only models and thus do not use this \texttt{fill-in-the-blank} scheme.} 
{As in \citet{llm_longtail}, we study few-shot settings and prompt LMs without any parameter updates, instead of fine-tuning them on QA datasets such as in \citet{roberts2020much}. }

\vspace{.1cm}
\noindent {\bf Metrics: accuracy.}
We mark a prediction as correct if any substring of the prediction is an exact match of any of the gold answers.


\subsection{Dimensions of Analysis}

We hypothesize that factual knowledge that is less frequently discussed on the web may not be well-memorized by LMs.
Previous research often uses the term frequency of object entities in pretraining corpora to understand memorization~\cite{fevry2020entities,llm_longtail, Razeghi2022ImpactOP}. 
{Instead, we investigate whether it's possible to predict memorization based on the input information only, and then apply the findings for modeling improvements, unlike prior analyses.}
Therefore, our work focuses on the other two variables in a factual knowledge triple: the subject entity and the relationship type. 
 


\vspace{.1cm}
\noindent {\bf Subject entity popularity.}
We use the popularity of the entities measured by Wikipedia monthly page views as a proxy for how frequently the entities are likely to be discussed on the web, instead of using the occurrence of entities or strings in the pretraining corpus~\cite{Carlini2022QuantifyingMA,llm_longtail,Razeghi2022ImpactOP}. 
{Calculating frequencies over large pretraining corpora requires massive computations to link entities over billions of tokens, or can result in noisy estimations.\footnote{Moreover, several recent models like GPT-3 do not release their pretraining corpora, and it is an open question whether the frequencies in pretraining corpora reflect the frequencies in their private corpora. } }
{Our initial studies show that this is much cheaper\footnote{We can get page views by calling Wikipedia API. } and aligns well with our intuition.}

\vspace{.1cm}
\noindent {\bf Relationship type.}
We also consider the relationship types as key factors for factual knowledge memorization. 
For example, even given the same combinations of the subject and object entities, model performance can depend on the relationship types;
relationship types widely discussed can be easier to be memorized, while types that are less discussed may not be memorized much. 

\subsection{Benchmarks}

\vspace{.1cm}
\noindent {\bf \ours.}
In our preliminary studies, we found that existing common open-domain QA datasets such as Natural Questions (NQ; \citealt{kwiatkowski-etal-2019-natural}) are often dominated by subject entities with high popularity, and it is often hard to identify relationship types due to diverse question surface forms.
To enable a fine-grained analysis of memorization based on the aforementioned analysis dimensions, we construct \ours, a new large-scale entity-centric open-domain QA dataset about entities with a wide variety of popularity, as shown in Figure~\ref{fig:pop_distr}.

To construct \ours, we randomly sample knowledge triples of 16 diverse relationship types from Wikidata and convert them into natural language questions, using a natural language template (depicted in Figure~\ref{fig:background}). 
We verbalize a knowledge triple \((S, R, O)\) into a question that involves substituting the subject \(S\) into a template manually written for the relationship type \(R\). 
The full list of templates is found in Table~\ref{tab:list_of_instructions_ours} of the Appendix. 
The set of acceptable answers to the question is the set of entities \(E\) such that \((S, R, E)\) exists in the knowledge graph. 
We tried various templates and found that the results were fairly robust to the templates. 
Since \ours\ is grounded to a knowledge base, links to Wikidata entities allow for reliable analysis of popularity and relationship types.

\vspace{.1cm}
\noindent {\bf EntityQuestions.}
We test on another popular open-domain QA dataset, EntityQuestions~\cite{sciavolino2021simple}, which also covers a long-tail entity distribution. 
They use Wikipedia hyperlink counts as a proxy of the frequency of entities and sample knowledge triples from WikiData, from the frequency distributions.
Unlike \ours, EntityQuestions doesn't provide entity annotations, so we only use 82\% of the questions, where the mention of the subject entity has a unique match with a Wikidata entity.

\section{Memorization Depends on Popularity and Relationship Type  }
\label{sec:analysis_parametric}
We evaluate a range of LMs with varying numbers of parameters, to quantify how much factual knowledge they memorize and how different factors 
affect those memorization behaviors \ref{Q1}. 

\subsection{Experimental Setup}
\paragraph{Models.}
We evaluate ten models with a varying scale of model size: OPT (\citealt{zhang2022opt}; 1.3, 2.7, 6.7, and 13 billion), GPT-Neo (\citealt{black2022gpt}; 1.3, 2.7, 6, and 20 billion), and GPT-3 (\citealt{brown2020language}; \texttt{davinci-002}, \texttt{davinci-003}) on our benchmark without any fine-tuning.\footnote{We did not explore widely-used encoder-decoder models such as T5, as their supervised pretraining consists of QA.}  

\paragraph{Instructions and demonstrations.}
We use a simple template ``\texttt{Q:~<question>~A:}'' to format all of our questions for generative prediction. More sophisticated instructions were attempted in preliminary experiments but they did not improve upon the simple template significantly enough to merit using them, especially given that they may overfit to the model. 
While we use zero-shot prompting for GPT-3 to reduce API costs,\footnote{Using 15-shot prompts for GPT-3 would cost upwards of \$3000 for the combination of vanilla, Contriever, BM25, and GenRead evaluations on \texttt{davinci-002} and \texttt{davinci-003}.} we use 15-shot prompting for all GPT-neo and OPT models. 


\begin{figure*}[t!]
    \centering
\includegraphics[width=\linewidth,trim=0cm 0.9cm 0cm 0cm]{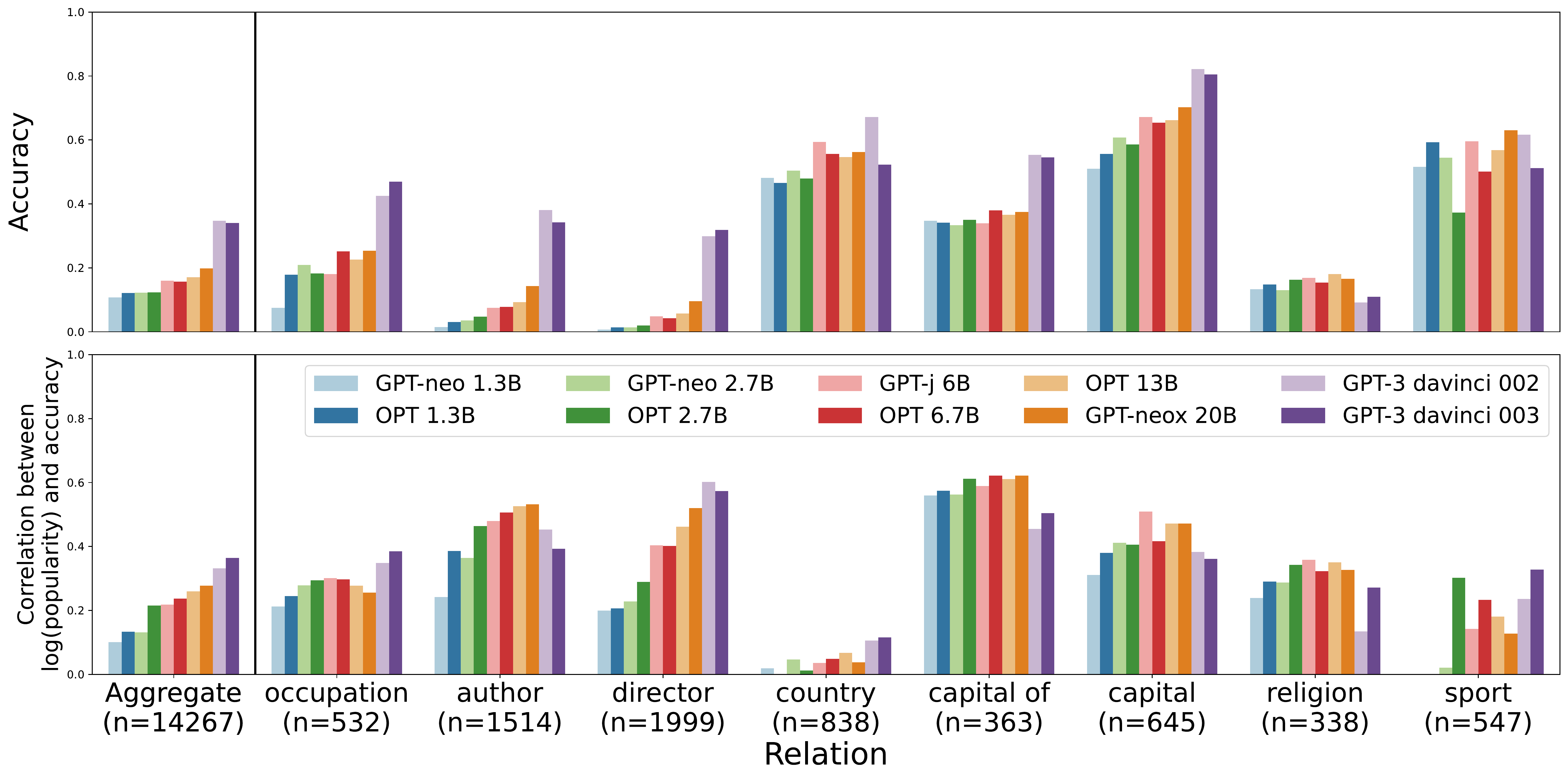}
    \caption{Per relationship type ($n =$ the number of questions) results on \ours{} by model, showing overall accuracy and the correlation between accuracy and log popularity. 
    We report the correlation between the question's subject entity log-popularity and the binary indicator of whether the question was answered correctly.
    We see that {\bf both subject entity popularity and relationship type are strong predictors of memorization across models.}
    The correlation with popularity {\bf exists across relationship types and is stronger for larger LMs}. 
    We show a representative subset of relationship types and the 
    complete results are in Figures~\ref{fig:appendix_accuracy_breakdown} and \ref{fig:appendix_correlation_breakdown} in Appendix~\ref{app_sec:lm_results}{, including results on EntityQuestions.}}
    \label{fig:relationship__all}
\end{figure*}
\subsection{Results}

\paragraph{Overall model performance.}
The top left column of Figure~\ref{fig:relationship__all} illustrates the overall performance on \ours. 
As shown, even without using in-context examples, larger LMs exhibit reasonable performance: GPT-3 achieves 35\% accuracy, and GPT-Neo 20B achieves 25\% accuracy.
This indicates that large LMs memorize factual knowledge in their parameters to some extent. 
This section examines which types of knowledge are better memorized and what factors influence memorization.

\paragraph{Subject entity popularity predicts memorization.}
Figure~\ref{fig:relationship__all} (bottom) shows that there is a positive correlation between subject entity popularity and models' accuracy for almost all relationship types. 
This supports our hypothesis that subject entity popularity can be a reliable indicator of LMs' factual knowledge memorization. 
In general, the correlations between subject entity popularity and accuracy are stronger for larger LMs; GPT-3 003 shows the highest positive correlation (roughly 0.4) while GPT-Neo-1.3B shows relatively weak positive correlations (approximately 0.1). 

\paragraph{Relationship types affects memorization.}
We find that models have a higher average performance for some relationship types than for others. 
While this is evidence that factual knowledge of some relationship types are more easily memorized than others, we also observe that questions of certain relationship types can be easily {\it guessed} without memorizing the knowledge triple. 
Specifically, certain relationship types (e.g., nationalities) allow models to exploit surface-level artifacts in subject entity names~\cite{poerner-etal-2020-e,cao-etal-2021-knowledgeable}. 
Additionally, models often output the most dominant answer entities for questions about relationship types with fewer answer entities (e.g., red for the color relationship type). 
In Figure~\ref{fig:relationship__all}, relationships with lower correlation (e.g., country, sport) often shows higher accuracy, indicating that on those relationship types, models may exploit surface-level clues. 
On the other hand, for relationship types with relatively low accuracy (e.g., occupation, author, director), larger LMs often show a high correlation. 
Further details are in Appendix~\ref{app_sec:lm_results}.

\paragraph{{Scaling may not help with tail knowledge.}}
As seen in the left column of Figure~\ref{fig:relationship__all}, there are clear overall performance improvements with scale on the \ours{} dataset. 
However, Figure~\ref{fig:lm_scale} shows that on both \ours~and EntityQuestions, most of scaling's positive effect on parametric knowledge comes from questions with high popularity. 
Specifically, for the questions about the entities whose $\log_{10}{(\rm popularity)}$ is larger than 4, there is an improvement in accuracy as model size increases (red and yellow lines), while performance on questions with lower popularity remains relatively constant (blue and green lines). 
For the 4,000 least popular questions, GPT-Neo 6B, 20B, and GPT-3 \texttt{davinci-003} have 15\%, 16\%, and 19\% accuracy, respectively.

 \begin{figure}[b!]
\centering
  \includegraphics[width=0.99\linewidth,keepaspectratio]{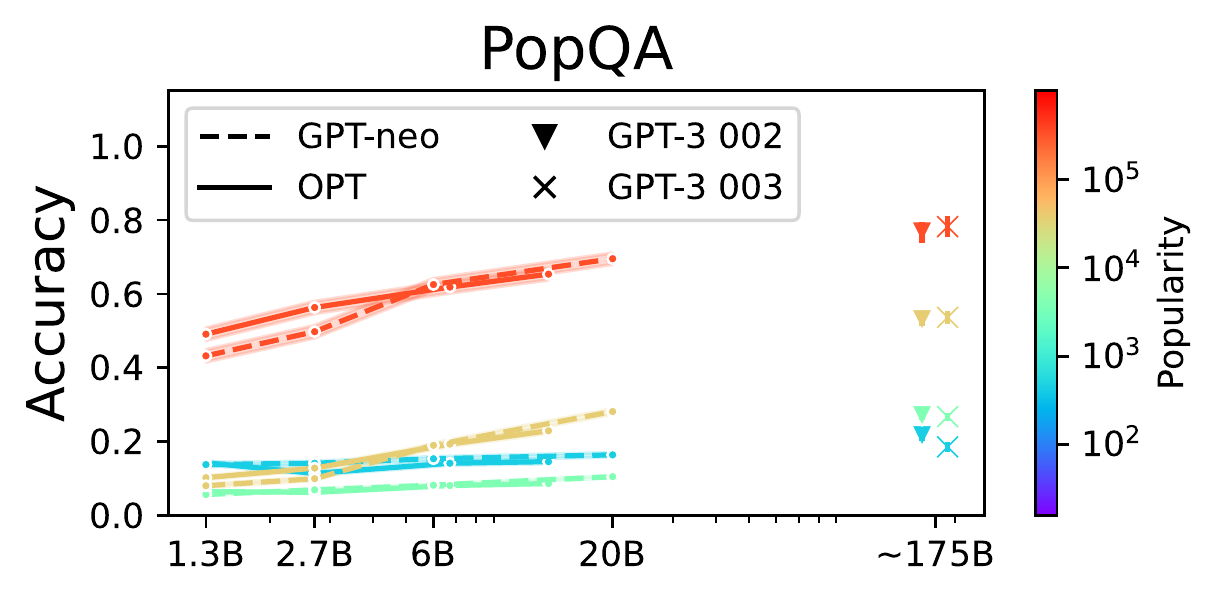}
  \includegraphics[width=0.99\linewidth,keepaspectratio]{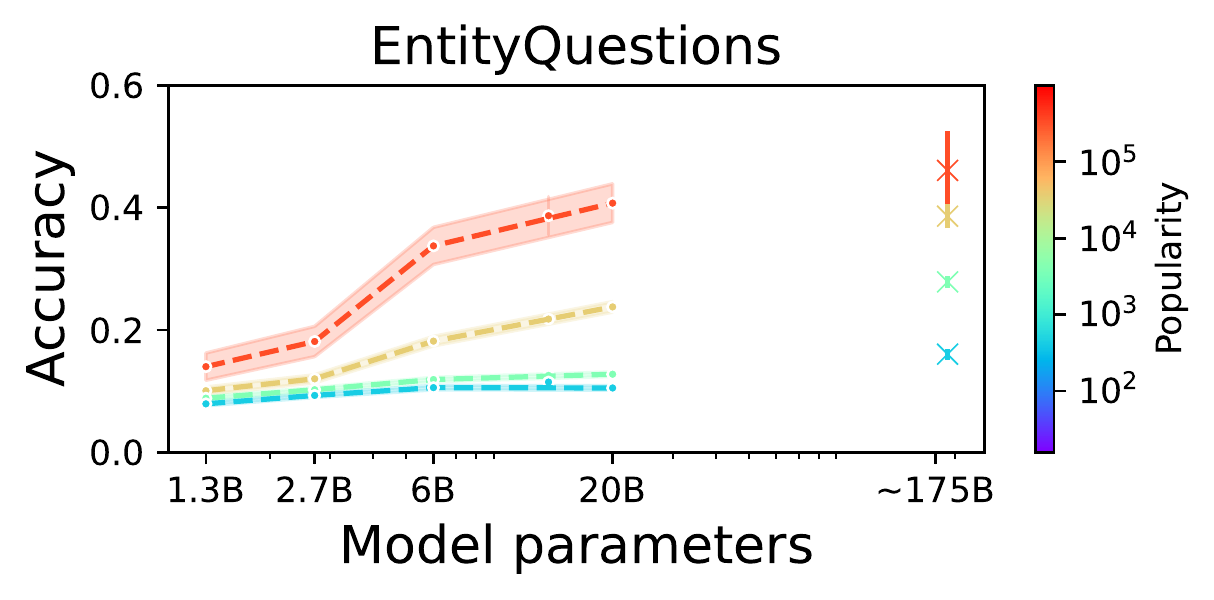}
  \captionsetup{width=0.95\linewidth}
\caption{\ours \ scaling results, broken down by question popularity level. 
{\bf Scaling mostly improves memorization of more popular factual knowledge.}
Error bars are 95\% confidence intervals.}
\label{fig:lm_scale}
\end{figure}
This somewhat dampens prior works' findings that scaling up models significantly improves their factual knowledge memorization~\cite{roberts2020much,llm_longtail}.
We hypothesize that this is because their evaluations are often conducted on QA datasets with popular entities.
\footnotetext{30 \ours\ 
 and 26 EntityQuestions questions had popularity less than the smallest popularity bin, and are excluded to avoid showing results for small sample sizes. }
{In sum, scaling lowers the threshold of popularity for knowledge to be reliably memorized, but is not projected to move the threshold far into the long tail for practical model scales.}

\paragraph{Relationship type results breakdown.}
Figure~\ref{fig:relationship_breakdown} provides a closer look at the relationship between popularity, accuracy, and relationship type; it shows model accuracy over the popularity distributions for director and country.  
For the first two types, we can see a clear positive trend between popularity and accuracy across models, and as the model size gets larger, the LMs memorize more. 
On the other hand, in the ``country'' relationship type, no models show trends, while overall the accuracy is high, indicating the LMs often exploit artifacts to answer less popular questions. 
We show example models' predictions in Appendix Section~\ref{sec:qualitatitve_results}.
\section{Non-parametric Memory Complements Parametric Memory}
\label{sec:analysis_non_parametric}
Our analysis indicates that even the current state-of-the-art LMs struggle with less popular subjects or certain relationship types, and increasing the model size does not lead to further performance improvements.
In light of this, we extend our analysis to non-parametric sources of knowledge, as outlined in Section~\ref{Q2}. Specifically, we investigate the effectiveness of retrieval-augmented LMs~\cite{borgeaud2021improving,lewis2020retrieval}, which leverage non-parametric memories (i.e., retrieved text) to improve performance.

\subsection{Experimental Setup}
\noindent {\bf Augmenting input.} 
In this work, we try a simple retrieval-augmented LM approach, where we run an off-the-shelf retrieval system off-line to retrieve context from Wikipedia relevant to a question,\footnote{We use Wikipedia dump from December 2018.} and then we concatenate the retrieved context with the original question. 
Although increasing the context size often leads to performance gains~\cite{izacard2020leveraging,asai2021evidentiality}, we only use the top one retrieved paragraph for simplicity. 

\begin{figure}[t!]
    \includegraphics[width=\linewidth]{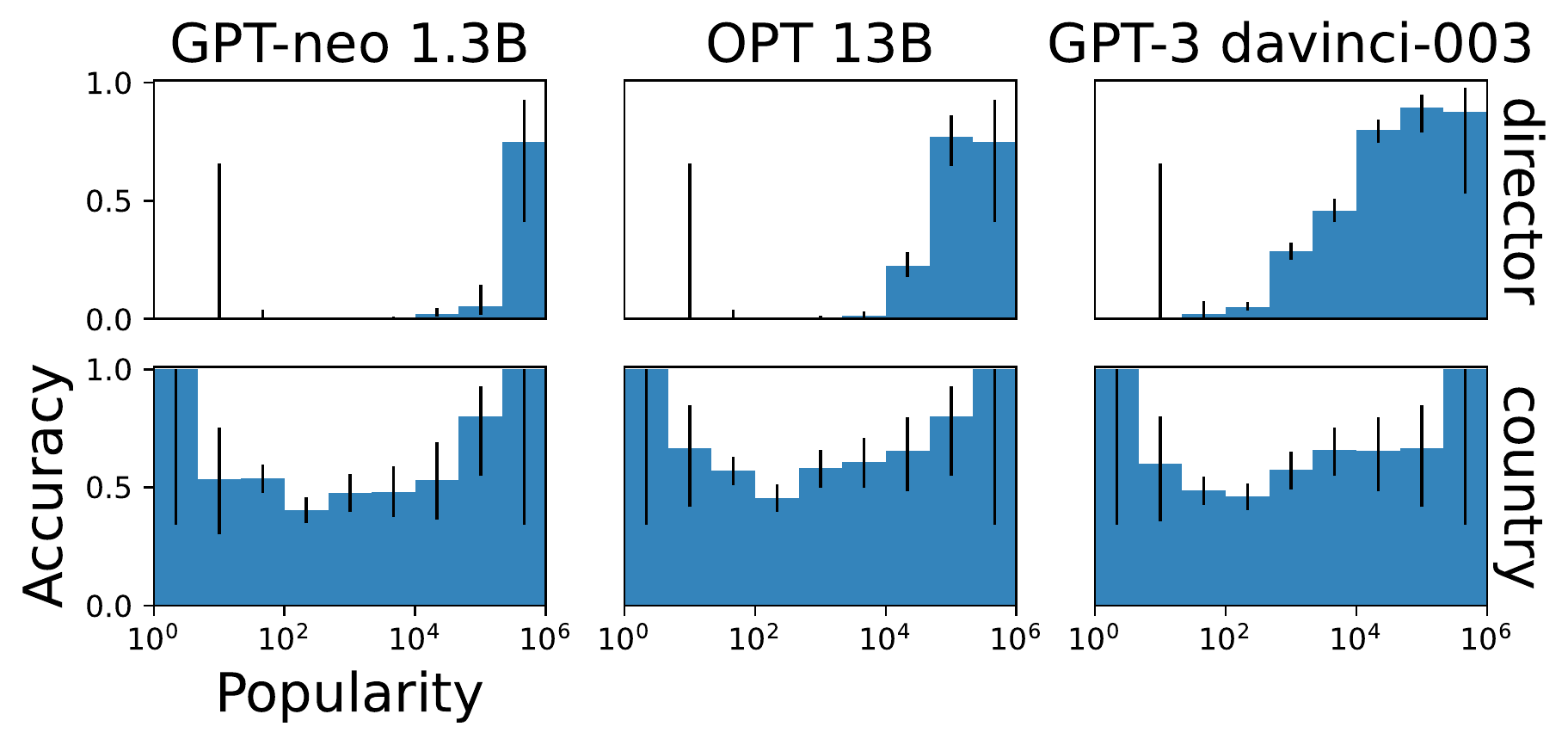}
    \caption{
Memorization versus popularity for three models and the relationship types with the largest and smallest correlations. 
Within a relationship type, generally, there is a {\bf monotonically increasing link between popularity and performance}, except for ``country''. Error bars show Wilson 95\% confidence intervals.
} 
\label{fig:relationship_breakdown}
\end{figure}
\vspace{.15cm}
\noindent {\bf Retrieval models.} 
We use two widely-used retrieval systems: {\bf BM25}~\cite{robertson2009probabilistic} and {\bf Contriever}~\cite{izacard2021towards}. 
BM25 is a static term-based retriever without training, while Contriever is pretrained on large unlabeled corpora, followed by fine-tuning on MS MARCO~\cite{bajaj2016ms}.
We also experiment with a \emph{parametric} augmentation method, {\bf GenRead}~\cite{yu2022generate}, which prompts LMs to generate rather than retrieve a contextual document to answer a question. We use the ten LMs in Section~\ref{sec:analysis_parametric}, resulting in 40 LMs and retrieval-augmented LMs.
\begin{figure}[ht!]
    \centering
    \includegraphics[width=0.99\linewidth]{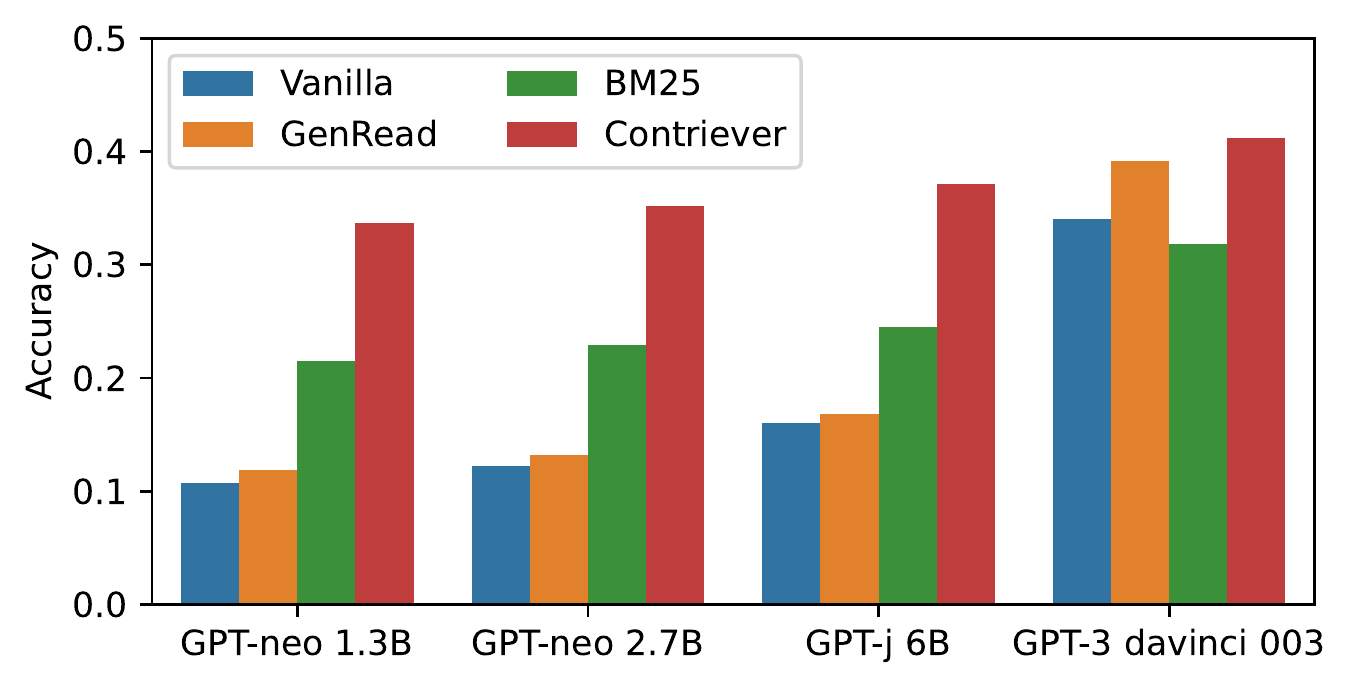}
    \caption{\ours~accuracy of LMs augmented with BM25, Contriever, GenRead, and unassisted (vanilla). 
    {\bf Retrieving non-parametric memories significantly improves the performance of smaller models. }
    {Complete results on \ours\ are found in Figure~\ref{fig:appendix_acc_by_model}. EntityQuestions results are in Figure~\ref{fig:appendix_EQ_acc_by_model} of the Appendix. }}
    \label{fig:acc_by_model}
\end{figure}

\subsection{Results}

\paragraph{Retrieval largely improves performance.} 
Figure~\ref{fig:acc_by_model} shows that augmenting LMs with non-parametric memories significantly outperforms unassisted vanilla LMs. 
A much smaller LM (e.g., GPT-Neo 2.7B) augmented by the Contriever retrieval results outperforms vanilla GPT-3. 
Large LMs such as GPT-3 also enjoy the benefits of non-parametric memories. Contriever gives 7\% accuracy gains on top of GPT-3 \texttt{davinci-003}. 
GenRead shows little-to-no performance improvement over vanilla parametric knowledge for smaller models, while the technique shows sizeable gains for GPT-3, especially \texttt{davinci-003}. 
{In addition to its limited effectiveness with smaller LMs, }
GenRead has potentially prohibitive inference time costs, with GPT-NeoX 20B taking 70 seconds per query.

\begin{figure}[ht!]
    \centering
    \includegraphics[width=0.95\linewidth]{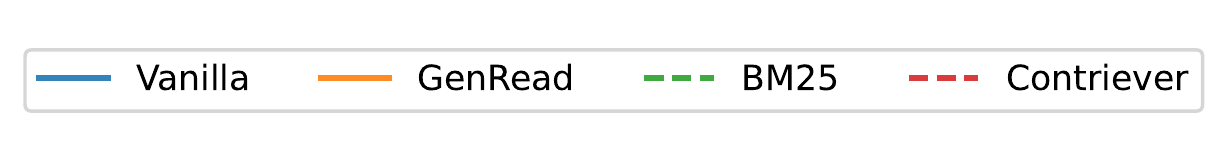}
    \includegraphics[width=0.95\linewidth]{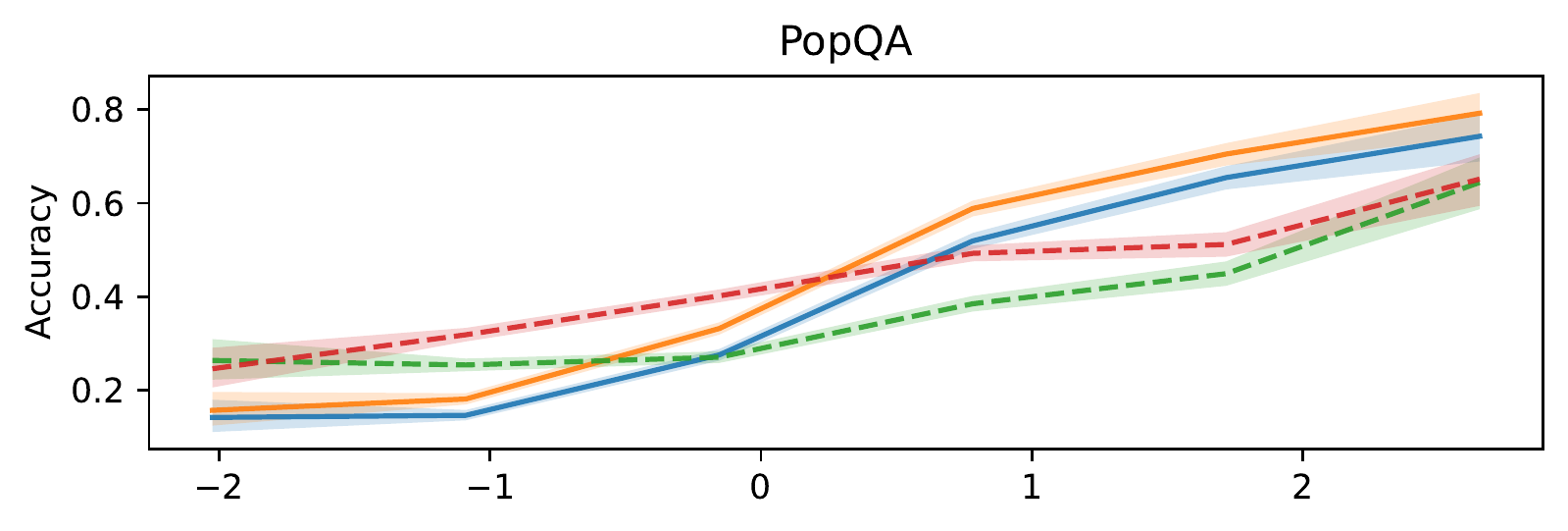}
    \includegraphics[width=0.95\linewidth]{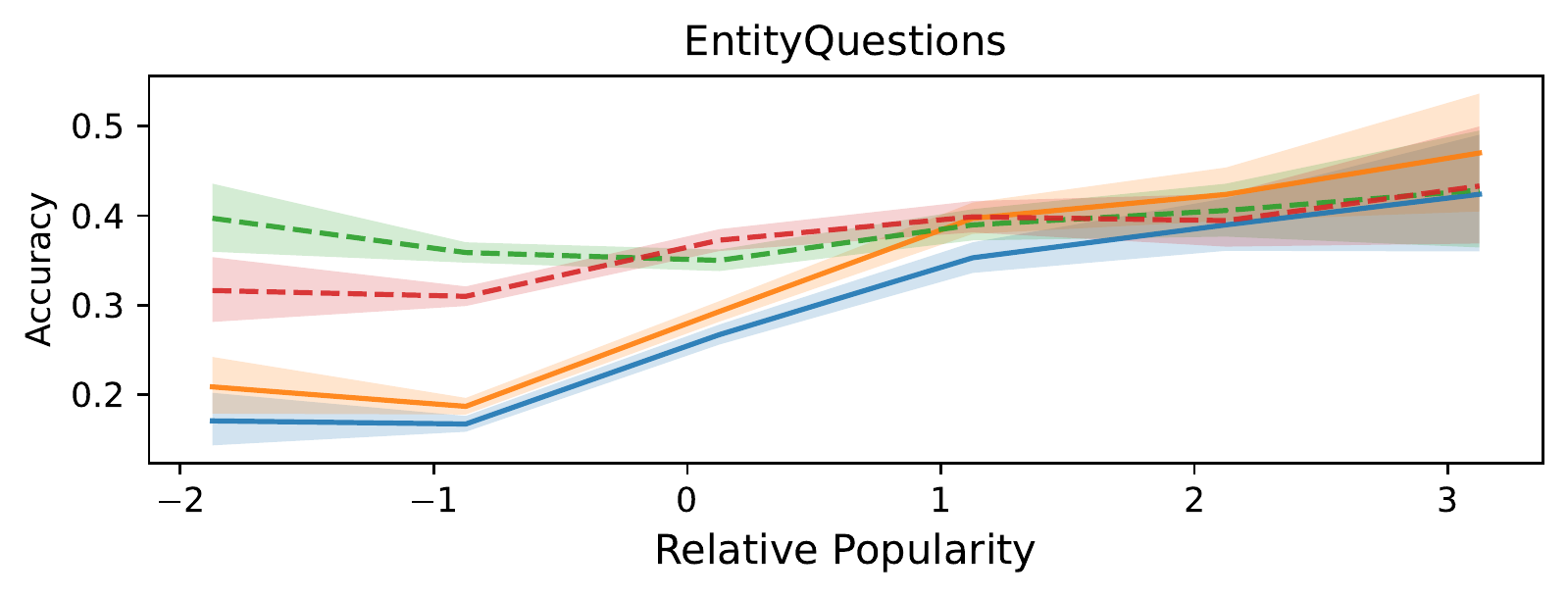}
    
    \caption{ 
    GPT-3 \texttt{davinci-003} accuracy versus relative popularity 
    (how popular a question is relative to other questions of its relationship type). \textbf{Retrieval-augmented LMs (dashed) outperform LMs' parametric memory (solid) for less popular entities, while parametric memory is competitive for more popular entities.} 
    Relative popularity is defined as the log-popularity of a question, normalized by the mean and standard deviation of log-popularity for the question's relationship type (smaller for less popular entities).\footnotemark\ Figure~\ref{fig:appendix_correlation_breakdown} shows per-relationship results. 
    }
    \label{fig:retrieval_how_helps}
\end{figure}
\footnotetext{Error bars show Wilson 95\% confidence intervals. Bins with less than 40 samples have been excluded to avoid showing results with exceedingly wide errorbars.}

\paragraph{Non-parametric memories are effective for less popular facts.} 
How does retrieval augmentation lead to such significant improvements? 
Figure~\ref{fig:retrieval_how_helps} shows the relationship between the entity popularity and models' QA performance. 
It can be seen that retrieval-augmented LMs guided by Contriever or BM25 have a clear advantage over unassisted vanilla LMs, especially on less popular entities, resulting in a significant performance gain. 
Overall, Contriever-guided LMs outperform BM25-based ones on \ours, while the BM25-based models perform better on the least popular entities, consistent with the findings from \citet{sciavolino2021simple}.
On the other hand, for more popular entities, parametric knowledge shows equal or higher accuracy, indicating that the state-of-the-art LMs have already memorized the answers, and augmenting input with retrieved-context doesn't help much or even hurts the performance. 
Interestingly, GenRead generally outperforms vanilla LMs despite relying on LMs' parametric memory. This demonstrates the effectiveness of elicitive prompting~\cite{wei2022chain,sun2022recitation}
as observed in prior work. 
However, like vanilla LMs, GenRead shows low performance on less popular entities. 

\begin{table}[t!]
\centering
\footnotesize
\begin{tabular}{l|ll}\toprule
& \multicolumn{2}{c}{Contriever-augmented LM} \\ &  succeeded & failed \\ \midrule
LM succeeded & 0.83 (24\%) & 0.14 (10\%)\\ 
LM failed & 0.88 (17\%) & 0.11 (49\%) \\
\bottomrule
\end{tabular}
    \caption{ The recall@1 of Contriever for questions that GPT-3 \texttt{davinci-003} answered correctly and incorrectly with and without retrieval on \ours.  The percent of questions falling in each category is shown in parentheses. \textbf{For 10\% of questions, retrieval is \emph{harmful} due to low-quality retrieved text (0.14 recall@1)}. 
    }
    \label{tab:four_quad}
\end{table}

\paragraph{Non-parametric memories can mislead LMs.} 
We conduct an in-depth analysis of why retrieval-augmented models suffer in more popular entities. 
We hypothesize that retrieval results may not always be correct or helpful, and can mislead LMs. 
{To test this hypothesis, we group the questions based on two axes: whether unassisted GPT-3 \texttt{davinci-003} predict correctly or not, and whether retrieval-augmented predictions are correct or not. For each of the four categories, we calculate recall@1 (whether a gold answer is included in the top 1 document; ~\citealt{karpukhin2020dense}). }

Table~\ref{tab:four_quad} shows recall@1 for each group with percentages of the questions falling into each of the categories.
For 10\% of questions, retrieval-augmentation causes the LM to incorrectly answer a question it could otherwise answer correctly.  
We found that on those questions, recall@1 is significantly lower than the overall recall@1 (0.14 vs 0.42 overall), indicating that failed retrieval can result in performance drops. Conversely, for the 17\% of questions for which retrieval causes the LM to correctly answer a question it would otherwise have failed to answer, the recall@1 is 0.88.
We include examples of both cases in Appendix Section~\ref{sec:qualitatitve_results}.

\section{Adaptive Retrieval: Using Retrieval Only Where It Helps}
\label{sec:adaptive_retrieval}
While incorporating non-parametric memories helps in long-tail distributions, powerful LMs have already memorized factual knowledge for popular entities, and retrieval augmentation can be harmful. 
As outlined in \ref{Q3}, can we achieve the best of both worlds?
We propose a simple-yet-effective method, Adaptive Retrieval, which decides when to retrieve passages only based on input query information and augments the input with retrieved non-parametric memories only when necessary. 
We show that this is not only more powerful than LMs or retrieval-augmented LMs always retrieving context, but also more efficient than the standard retrieval-augmented setup.

\subsection{Method}
Adaptive Retrieval is based on our findings: as the current best LMs have already memorized more popular knowledge, we can use retrieval only when they do not memorize the factual knowledge and thus need to find external non-parametric knowledge. 
In particular, we use retrieval for questions whose popularity is lower than a threshold ({\it popularity threshold}), and for more popular entities, do not use retrieval at all. 

Using a development set, the threshold is chosen to maximize the adaptive accuracy, which we define as the accuracy attained by taking the predictions of the retrieval-augmented system for questions below the popularity threshold and the predictions based on parametric knowledge for the rest.
We 
determine the popularity threshold independently for each relationship type.


\begin{figure}[ht!]
    \centering
    \includegraphics[width=0.99\linewidth]{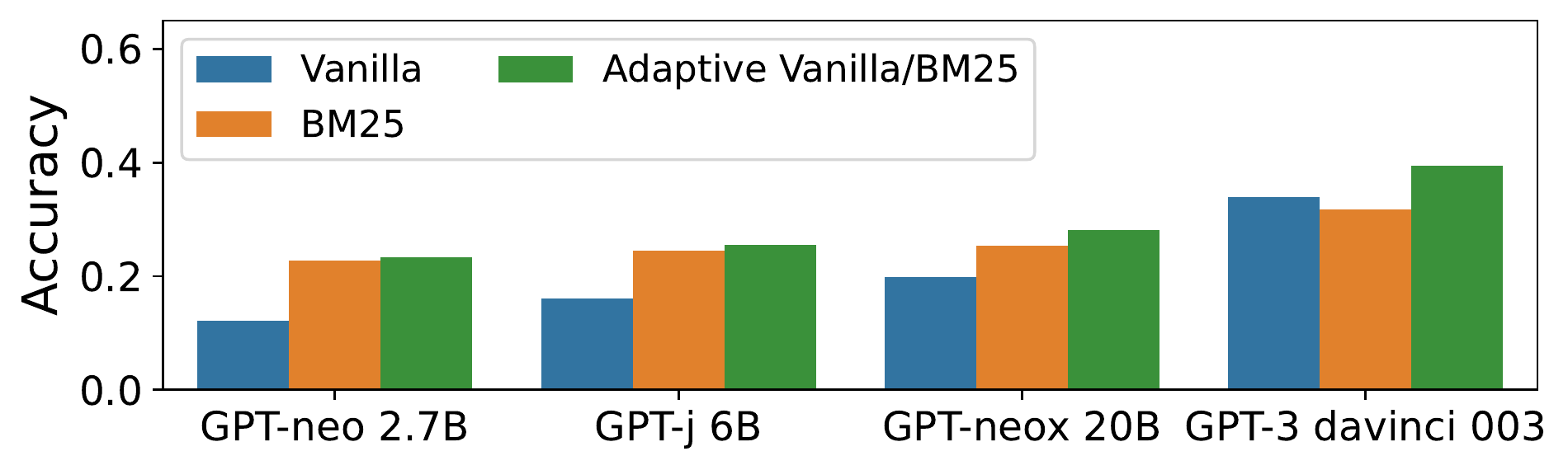}
    \caption{\ours\ performance of GPT-neo models and GPT3 \texttt{davinci-003}, with different retrieval methods. 
    \textbf{Adaptive Retrieval robustly outperforms approaches that always retrieve, especially for larger LMs}. 
    }
    \label{fig:popQA_adaptive}
\end{figure}
\subsection{Results}
\paragraph{Adaptive Retrieval improves performance.}
Figure~\ref{fig:popQA_adaptive} shows the results when we adaptively retrieve non-parametric memories based on the per-relationship type thresholds. 
We can see that adaptively retrieving non-parametric memories is effective for larger models.
The best performance on \ours~is using GPT-3 \texttt{davinci-003} adaptively with GenRead and Contriever, yielding 46.5\% accuracy, 5.3\% higher than any non-adaptive method.

\begin{figure}[ht!]
    \centering
    \includegraphics[width=0.99\linewidth, trim=0cm 0.5cm 0cm 0.3cm]{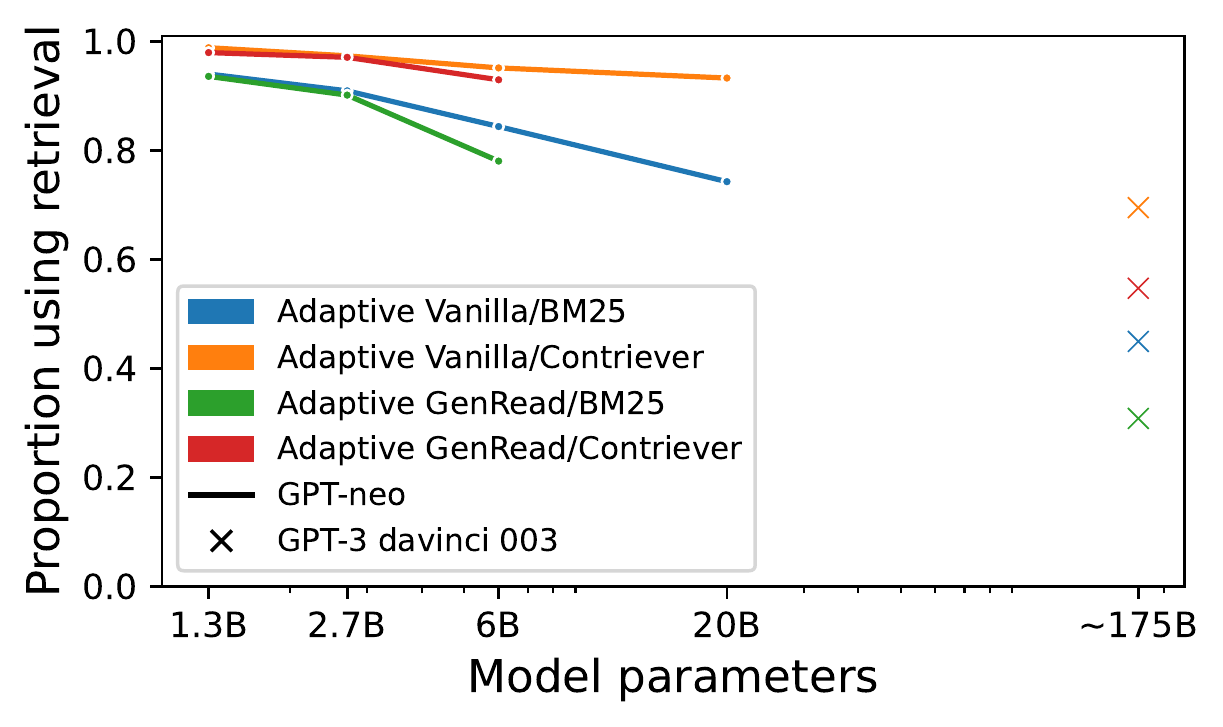}
    \caption{The proportion of questions for which various models use retrieval in the Adaptive Retrieval setup on \ours. When using Adaptive Retrieval, small models must still rely on non-parametric memory for most questions, while larger models have more reliable parametric memories enabling them to use retrieval less often.
    \label{fig:retr_savings_scale}}
\end{figure}

\paragraph{The threshold shifts with LM scale.}
While Adaptive Retrieval shows performance gains for larger models, smaller models do not realize the same benefits; as shown in Figure~\ref{fig:popQA_adaptive}, the performance gain from Adaptive Retrieval is much smaller when we use models smaller than 10 billion. Why does this happen? Figure~\ref{fig:retr_savings_scale} shows that smaller LMs almost always retrieve, indicating that there are not many questions for which small LMs' parametric knowledge is more reliable than non-parametric memory.
In contrast, large models typically retrieve much less. For example, GPT-3 \texttt{davinci-003} only retrieves for 40\% of questions when paired with BM25, and even the much smaller GPT-neox 20B does not retrieve documents on more than 20\% of the questions. {On EntityQuestions (Appendix Figure~\ref{fig:appendix_EQ_thresh_by_model}) all of the LMs retrieve much more, as the questions are mostly about less popular entities. }

\paragraph{Adaptive Retrieval reduces inference-time costs.}
We also found that Adaptive Retrieval improves efficiency; if we know we do not need to retrieve documents, we can skip retrieval components and the input length becomes shorter, which improves latency in both retrieval and language model components.
Figure~\ref{fig:adaptive_costs} shows the inference latency of GPT-J 6B and GPT-neox 20B, and API costs of GPT-3. 
Especially for larger LMs, concatenating retrieved context results in significantly increased latency (e.g., for GPT-J 6B, the inference time latency almost doubles). 
Adaptive retrieval enables reducing inference time up to 9\% from standard retrieval. 
{We also observe cost reduction on EntityQuestions, as shown in Figure~\ref{fig:EQ_adaptive}.}

\begin{figure}[t!]
    \centering
    \includegraphics[width=0.99\linewidth]{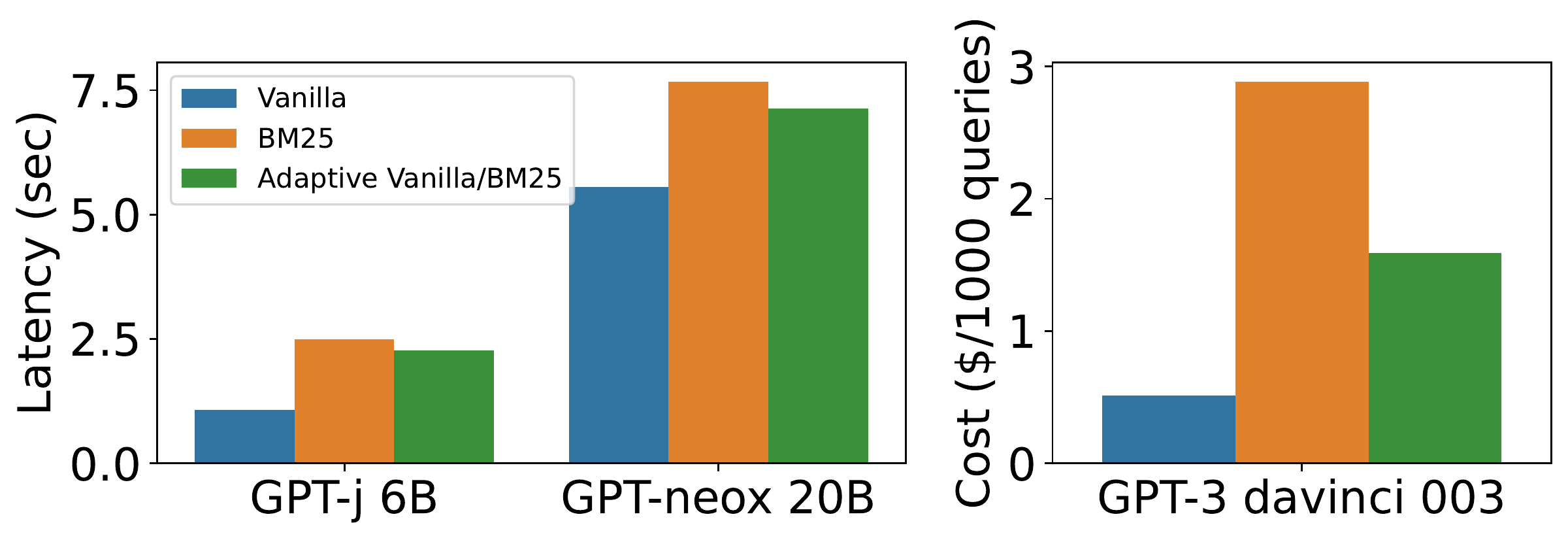}
    \caption{\ours\ latency for large GPT-neo models that were run on our machines, and API costs for GPT3. \bf{Adaptive retrieval reduces latency and API costs.}}
    \label{fig:adaptive_costs}
\end{figure}

\begin{figure}[t!]
    \centering
    \includegraphics[width=0.9\linewidth]{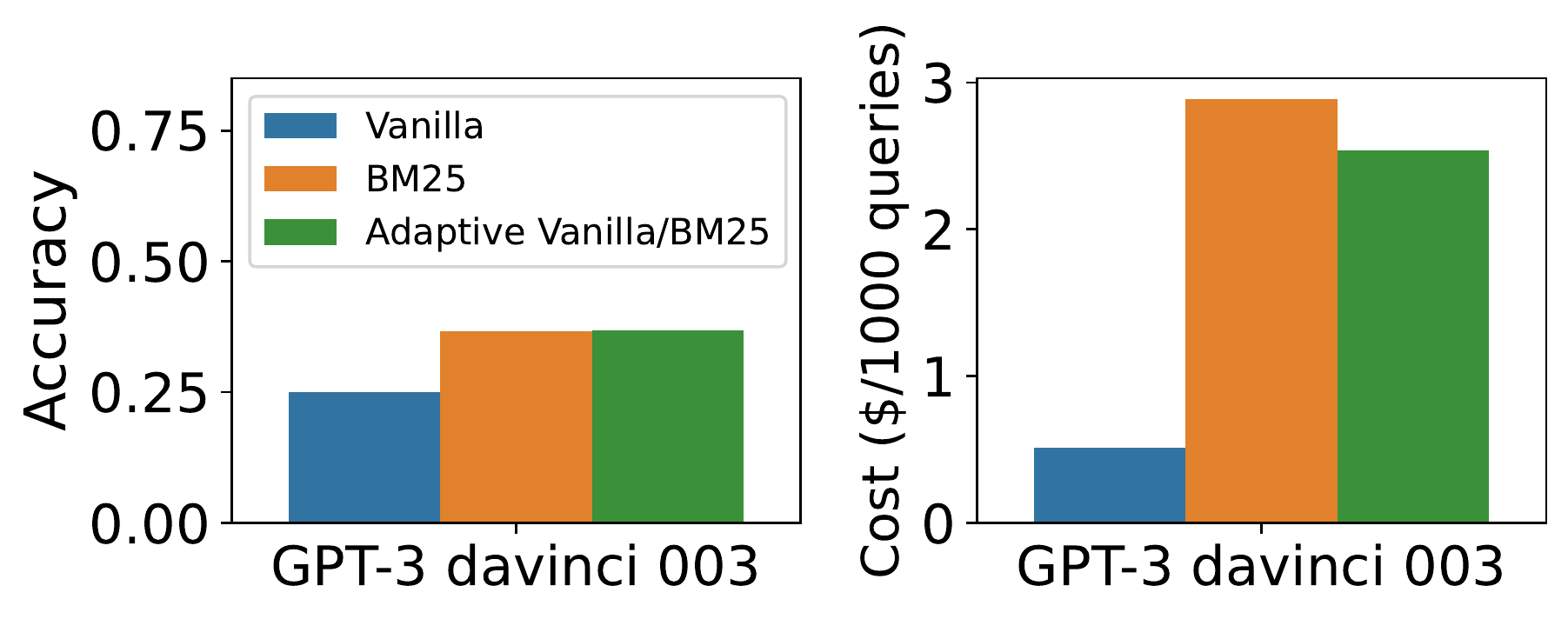}
    \caption{Accuracy and cost savings of Adaptive Retrieval for EntityQuestions. Despite EntityQuestions's lack of popular entities (see Figure~\ref{fig:pop_distr}), Adaptive Retrieval is able to reduce API costs by 15\% while maintaining equivalent performance to retrieval only.} \label{fig:EQ_adaptive}
\end{figure}
\section{Discussion and Conclusions}
This work conducts large-scale knowledge probing to examine the effectiveness and limitations of relying on LMs' parameters to memorize factual knowledge and to understand what factors affect factual knowledge memorization.
Our results show that memorization has a strong correlation with entity popularity and that scaling up models on long-tail distributions may only provide marginal improvements. 
We also demonstrate that non-parametric memories can greatly aid LMs on these long-tail distributions, but can also mislead LMs on questions about well-known entities, as powerful LMs have already memorized them in their parameters. 
Based on those findings, we devise simple-yet-effective Adaptive Retrieval, which only retrieves when necessary, using a heuristic based on entity popularity and relationship types. Our experimental results show that this method is not only more powerful than LMs or previous retrieval-augmented LMs but also more efficient. 

\section*{Limitations}
This work focuses on entity-centric factual knowledge and demonstrates that LMs' memorization is heavily affected by the popularity of the entities and the aspect of the entities being asked in the questions. 
It is important to emphasize that for running controlled experiments, we have relied on two synthetic datasets, and the extent to which our results apply to naturally occurring factual knowledge has not been firmly established. While we can be fairly confident about the relationship between scaling, retrieval, popularity, relationship type, and performance for the kinds of knowledge studied here, the effectiveness of Adaptive Retrieval will depend on many details of the question answering pipeline. 
Moreover, our work depends on a definition of popularity that is time-dependent and may not perfectly reflect how frequently entities are discussed on the web. Wikipedia page views are one possible definition of popularity for which we observe our results, and we invite others to improve upon it in future work. 
Further research can expand upon this simple approach, perhaps drawing on insights from~\citet{kadavath2022language} to improve the effectiveness of Adaptive Retrieval.

It is an open question if the same findings are applicable to other types of world knowledge such as commonsense. 
We conjecture that the concept of the subject topic (entity), as well as the aspect (relationship type), can be applied with some minor modifications, which future work can quantify memorization following our scheme.  

\section*{Ethical Considerations}
{
Recent work~\cite{huang2022large} shows that LMs memorize personal information available on the web, which has significant security issues. 
Our evaluation focuses on the memorization of general entity-centric knowledge, but our findings can be applicable to those areas. 
}
{Our findings suggest that LMs are likely to have less reliable knowledge of minority groups. \citet{parrish-etal-2022-bbq} established that models often rely on stereotypes to answer in uncertain cases, so our results indicate that LMs are likely to rely on stereotypes disproportionately for minority groups. Future work could investigate whether retrieval augmentation reduces bias in these cases.}

\section*{Acknowledgements}
We thank the UW NLP group members for their helpful discussions, and  Joongwon Kim, Wenya Wang, and Sean Welleck for their insightful feedback on this paper. 
This research was supported by NSF IIS-2044660, ONR N00014-18-1-2826, ONR MURI N00014- 18-1-2670, and Allen Distinguished Award. 
AM is funded by a Goldwater Scholarship and AA is funded by the IBM PhD Fellowship. 

\bibliography{anthology,custom}
\bibliographystyle{acl_natbib}

\clearpage
\appendix

\section*{Appendix}
\label{sec:appendix}
\section{Details of \ours~Constructions}
\label{app_sec:dataset}


\paragraph{List of the relationship types and templates.}
In this work, we use the following 16 relationship types, and the authors of this paper manually annotated templates to verbalize knowledge triple to natural language questions. 
We show the final list of the templates used to create \ours~in Table~\ref{tab:list_of_instructions_ours}. 

Figure~\ref{fig:pop_distr} shows the distribution of subject popularity of \ours and EntityQuestions versus the popular NQ benchmark. NQ may have multiple entities so the distribution of the least popular entity per question is shown. Subject entities from NQ were extracted using TagMe~\cite{Ferragina2010TAGMEOA} on the NQ-open development set with a score threshold of 0.22. TagMe returns the title of a Wikidata entity which can be directly used to find popularity.

\begin{table}[h!]
\footnotesize
\renewcommand{\arraystretch}{1.2}
\setlength{\tabcolsep}{2pt}
    \centering
    \begin{tabular}{lr}
\toprule
\textbf{Relationship } & \textbf{Template} \\\midrule
occupation  & What is \texttt{[subj]} 's occupation? \\
place of birth  & In what city was \texttt{[subj]} born? \\
genre  & What genre is \texttt{[subj]}?  \\
father  &  Who is the father of \texttt{[subj]} ?\\
country & In what country is \texttt{[subj]} ? \\
producer & Who was the producer of \texttt{[subj]} ? \\
director & Who was the director of \texttt{[subj]} ? \\
capital of & What is \texttt{[subj]}  the capital of? \\
screenwriter & Who was the screenwriter for \texttt{[subj]} ?\\
composer & Who was the composer of \texttt{[subj]} ? \\
color & What color is \texttt{[subj]} ? \\
religion & What is the religion of \texttt{[subj]} ? \\
sport & What sport does \texttt{[subj]}  play? \\
author & Who is the author of \texttt{[subj]} ? \\
mother & Who is the mother of \texttt{[subj]} ? \\
capital &What is the capital of \texttt{[subj]} ? \\
\bottomrule
 \end{tabular}
    \caption{Full list of the manually annotated templated used for \ours creations. \texttt{[subj]} denotes a placeholder for subject entities. }\label{tab:list_of_instructions_ours}
\end{table}

\paragraph{Knowledge triples sampling.} 
{In the construction of the \ours dataset, knowledge triples are sampled with higher weight given to more popular entities, otherwise, the distribution would be dominated by the tail and we would not have enough high-popularity entities to complete our analysis. 
Specifically, when considering whether to sample a particular knowledge triple, we include the knowledge triple if and only if \(f > \exp(8R - 6)\), where \(R \sim U(0, 1)\) is a unit uniform pseudo-random number and \(f\) is the exact match term frequency of the subject entity's aliases in an 800 MB random sample of C4. 
To increase diversity, once 2000 knowledge triples of a particular relation type have been sampled, they are no longer sampled.} 

\section{Experimental Details}

\paragraph{Computational resources and API costs.}
GPT-3 API usage totaled to \$275. We ran 14,282 questions through two GPT-3 \texttt{davinci} models using four different methods: vanilla experiments cost \$13 (\$0.46 per 1000 questions), Contriever-augmented experiments cost \$88 (\$3.08 per 1000 questions), BM25-augmented experiments cost \$81 (\$2.80 per 1000 questions), and GenRead experiments cost \$93 (\$3.25 per 1000 questions).


To run experiments using LMs larger than two billion parameters, we use a single V100 Volta GPU with 32GB GPU memories. We use int8bit~\cite{zeng2022glm} quantization with OPT 13 billion and GPT-Neo 20 billion models to make them fit our GPUs. In our preliminary experiments using GPT-Neo 6 billion, we did not observe a notable performance drop by using the quantization.  

\paragraph{Constructing few-shot contexts.} For \ours, we sample few-shot examples stratified by relationship type to diversify the samples: for each of the 15 relationship types other than the one in the test question, we sample one random question-answer pair to include in the context. For EntityQuestions, we take a simple random sample of 15 question-answer pairs because there are more than 16 relationship types. 

\paragraph{Details of deciding thresholds.} 
We 75\% of \ours to determine a popularity threshold for each relation type. Using brute force search, we select the threshold to maximize the adaptive accuracy, which we define as the accuracy attained by taking the predictions of the retrieval-augmented system for questions below the popularity threshold and the predictions based on parametric knowledge for the rest.

We then evaluate adaptive accuracy using the learned thresholds on the remaining 25\% of \ours, and repeat with 100 different random splits and take the mean to obtain the reported adaptive accuracy measurement.


\section{Detailed Results}
\subsection{LM results}
\label{app_sec:lm_results}

\paragraph{Full results of per-relationship type accuracy and correlation.}
Figure~\ref{fig:appendix_accuracy_breakdown} shows the full result of per-relationship type accuracy for all relationship types in \ours. 
Figure~\ref{fig:appendix_correlation_breakdown} shows the correlations for all relation types. Figures~\ref{fig:appendix_EQ_accuracy_breakdown} and~\ref{fig:appendix_EQ_correlation_breakdown} show the same results for the EntityQuestions dataset. 

\begin{figure}[t!]
    \centering
    \includegraphics[width=\linewidth]{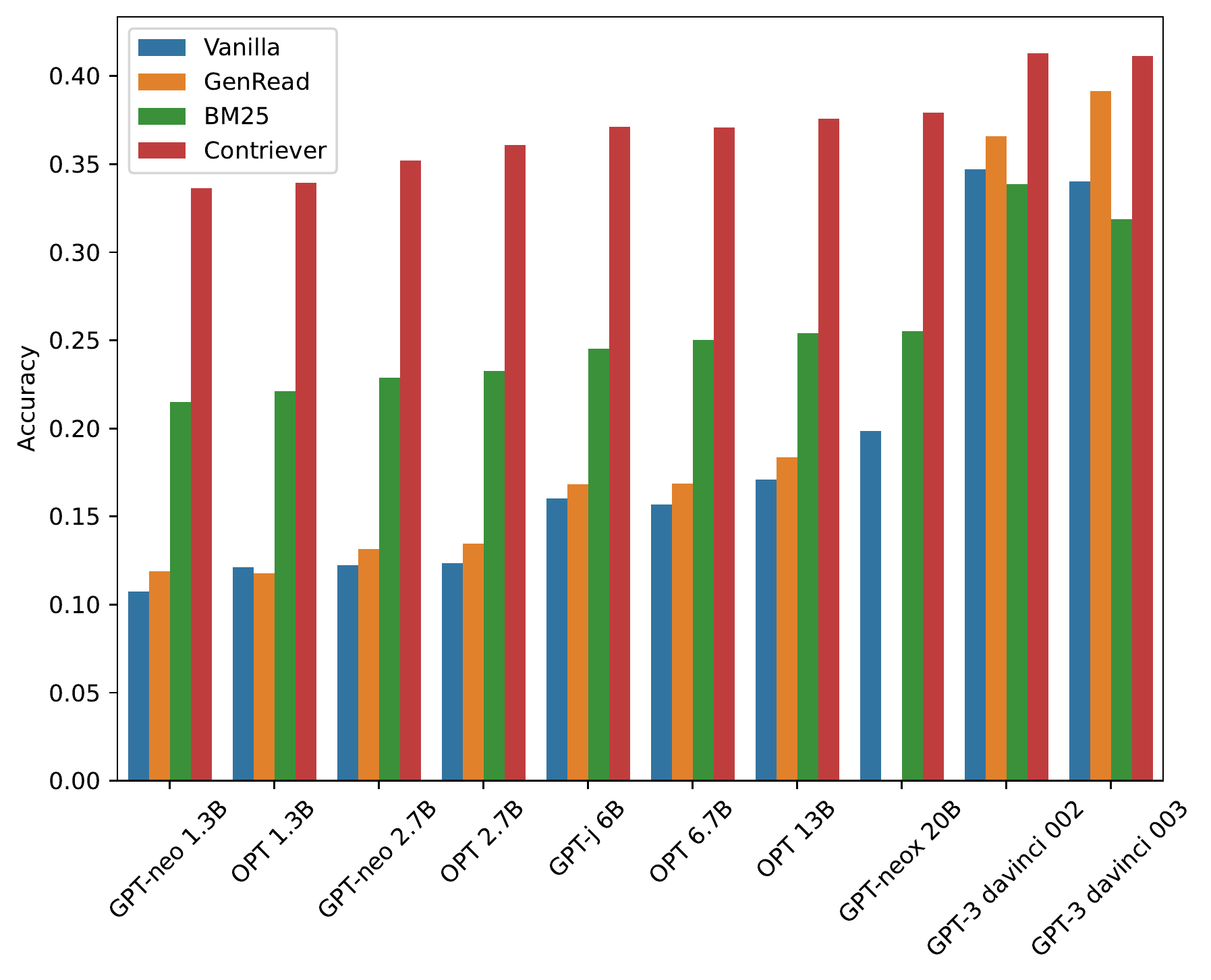}
    \caption{Accuracy by LMs and retrieval-augmented LMs on \ours. This is an extension of Figure~\ref{fig:acc_by_model} }
    \label{fig:appendix_acc_by_model}
\end{figure}

\begin{figure}[t!]
    \centering
    \includegraphics[width=\linewidth]{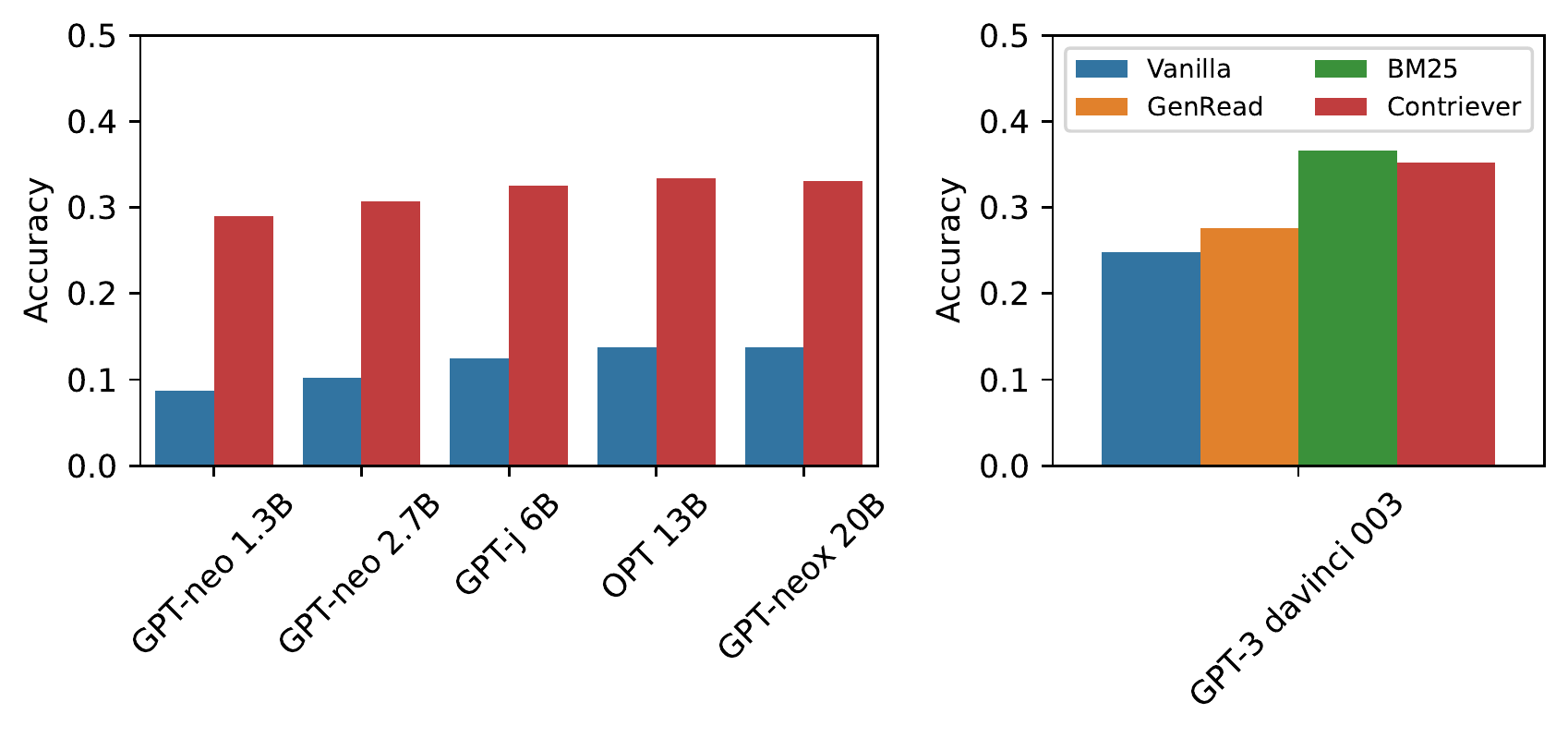}
    \caption{Accuracy by LMs and retrieval-augmented LMs on EntityQuestions.}
    \label{fig:appendix_EQ_acc_by_model}
\end{figure}

\begin{figure}[t!]
    \centering
    \includegraphics[width=\linewidth]{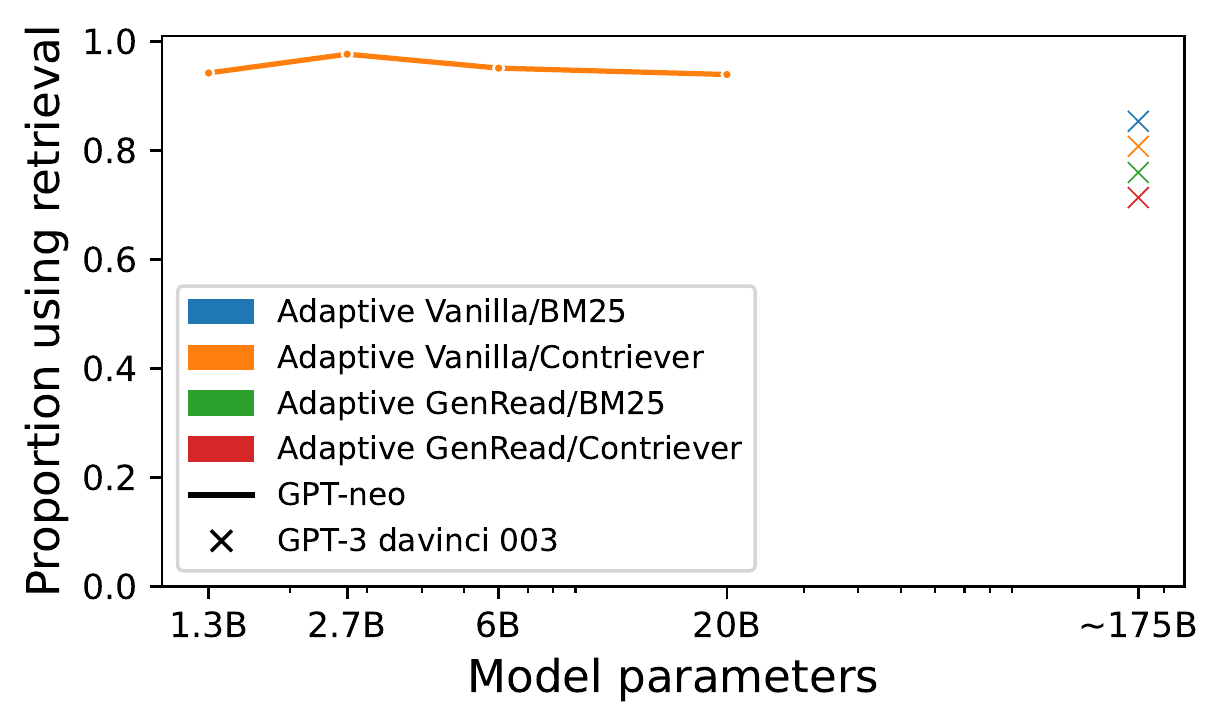}
    \caption{The proportion of questions for which Adaptive Retrieval uses retrieval versus model size for EntityQuestions.}
    \label{fig:appendix_EQ_thresh_by_model}
\end{figure}

\begin{figure*}
    \centering
    \includegraphics[width=0.9\textwidth]{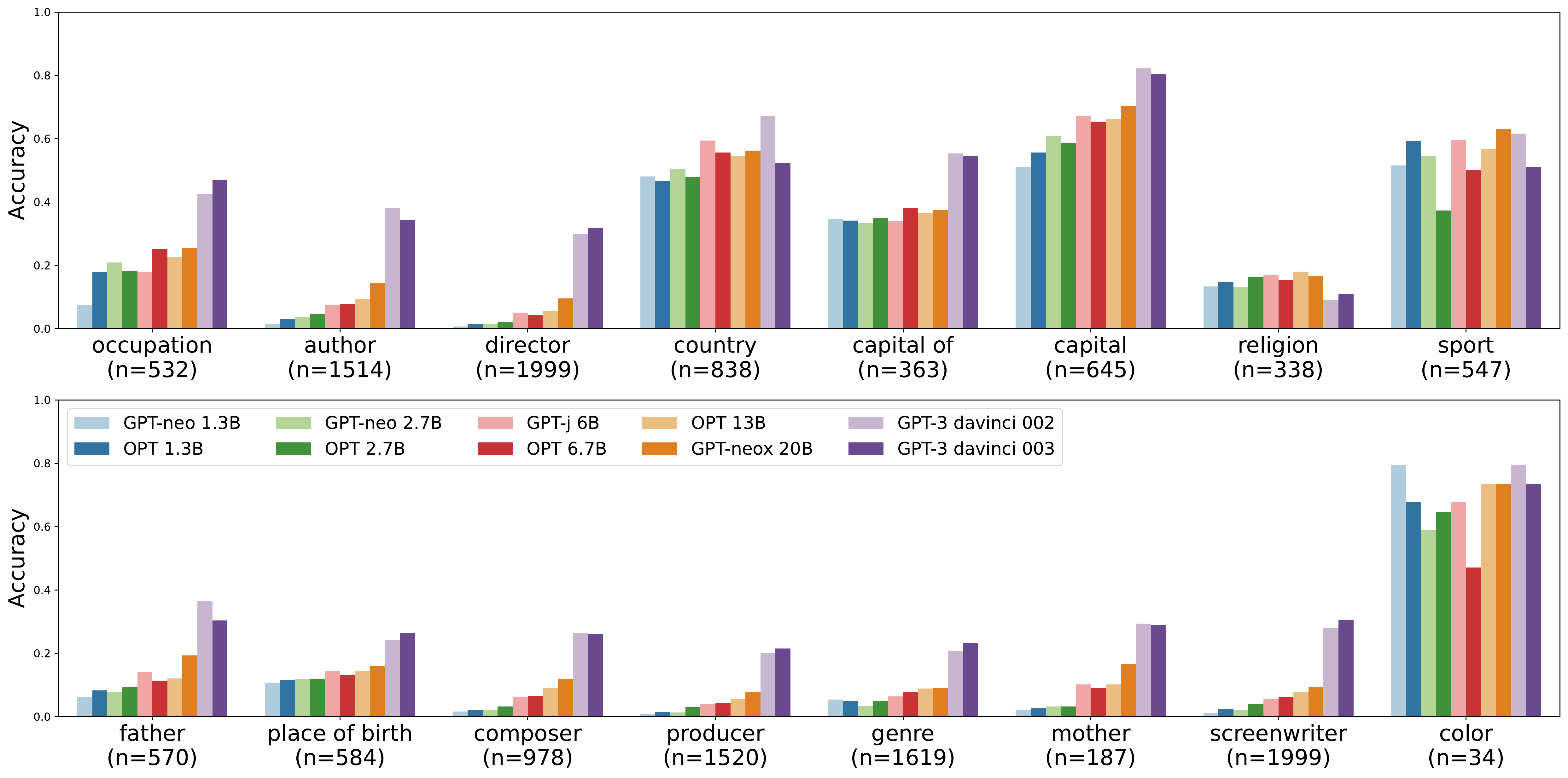}
    \caption{Accuracy on PopQA for all relationship types and models. This is an extension of Figure~\ref{fig:relationship__all}.}
    \label{fig:appendix_accuracy_breakdown}
\end{figure*}
\begin{figure*}
    \centering
    \includegraphics[width=0.9\textwidth]{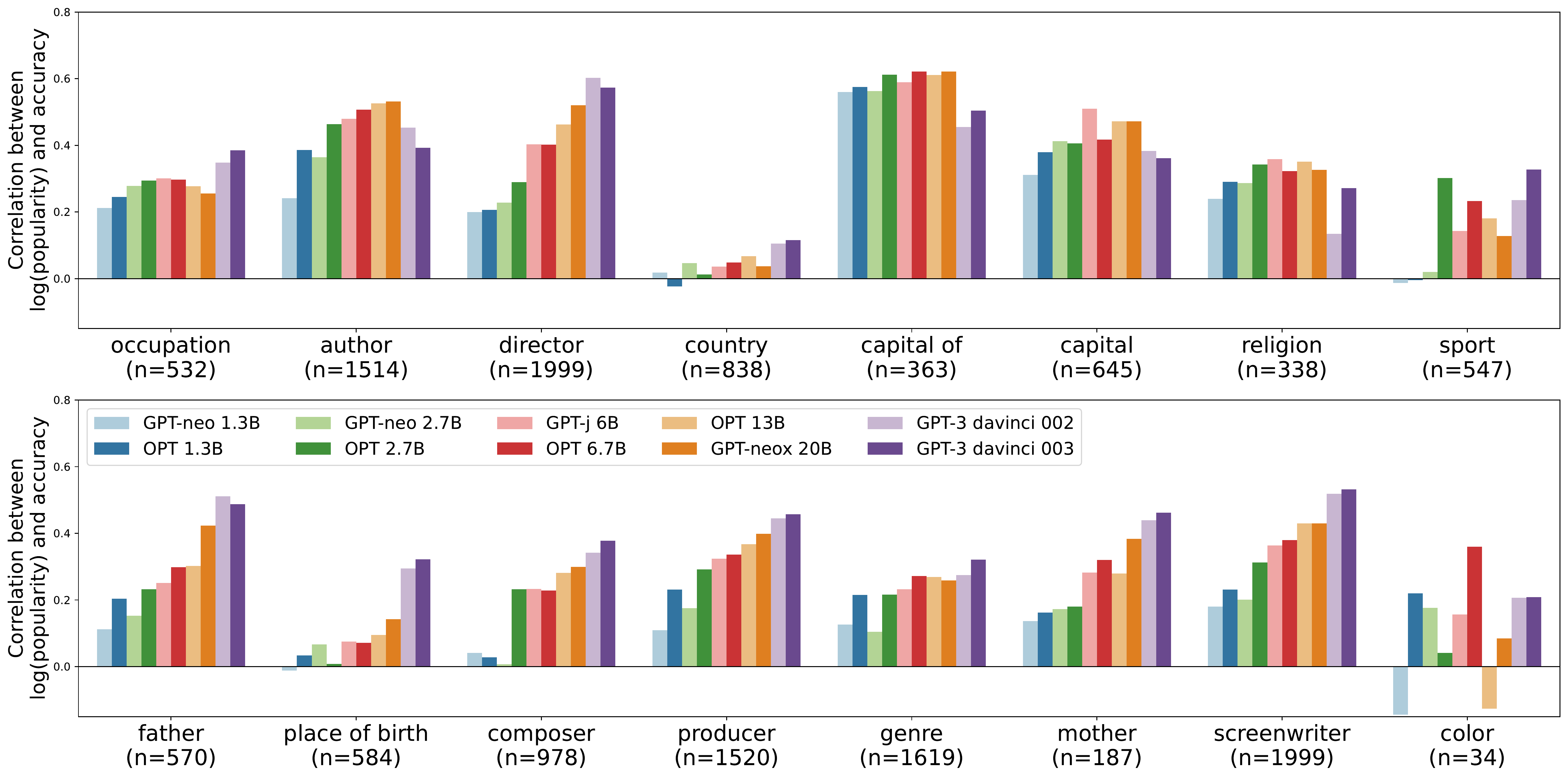}
    \caption{Correlations on PopQA for all relationship types and models. This is an extension of Figure~\ref{fig:relationship__all}.}
    \label{fig:appendix_correlation_breakdown}
\end{figure*}

\begin{figure*}
    \centering
    \includegraphics[width=0.9\textwidth]{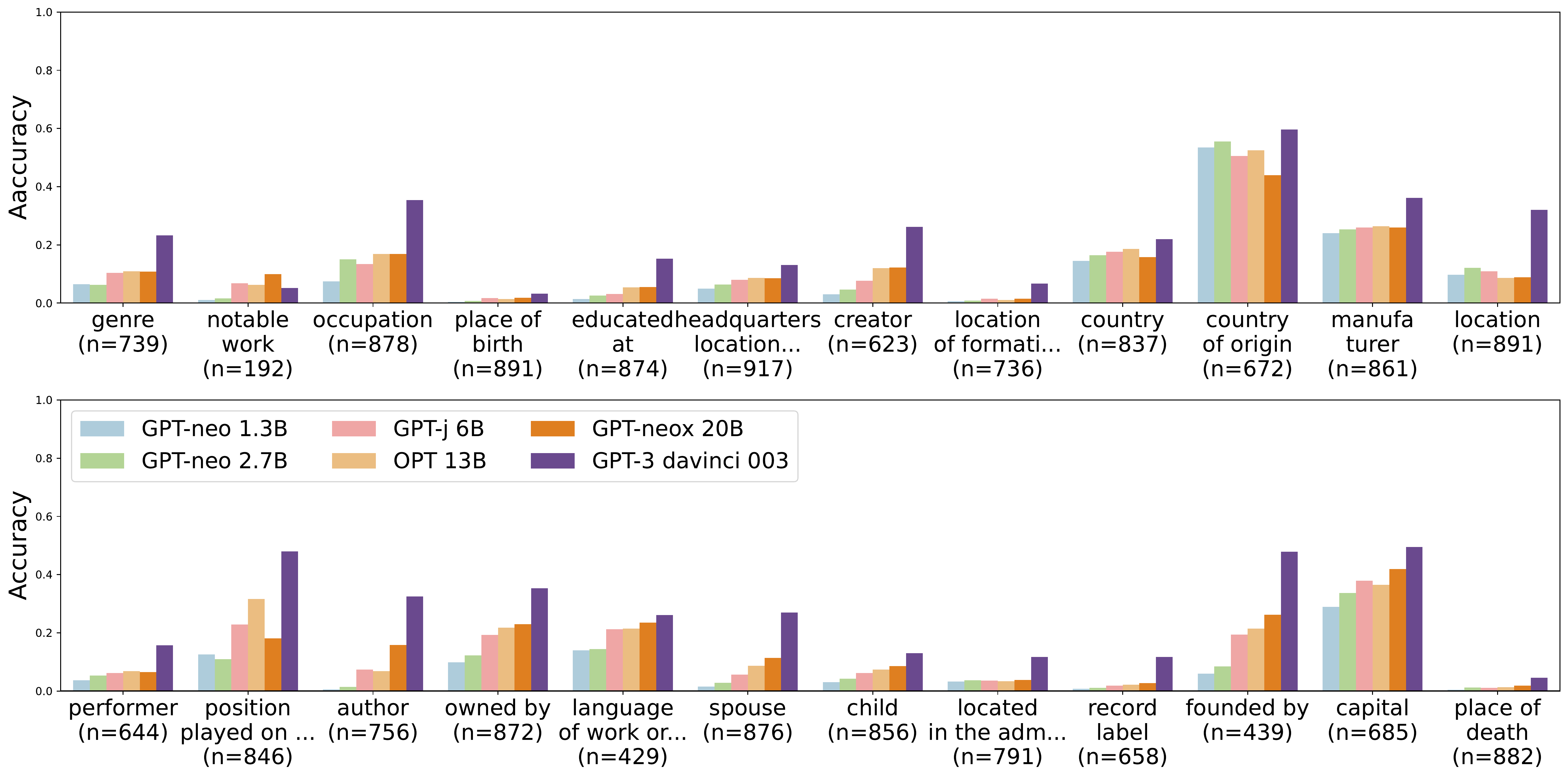}
    \caption{Accuracy on EntityQuestions for all relationship types and models.}
    \label{fig:appendix_EQ_correlation_breakdown}
\end{figure*}

\begin{figure*}
    \centering
    \includegraphics[width=0.9\textwidth]{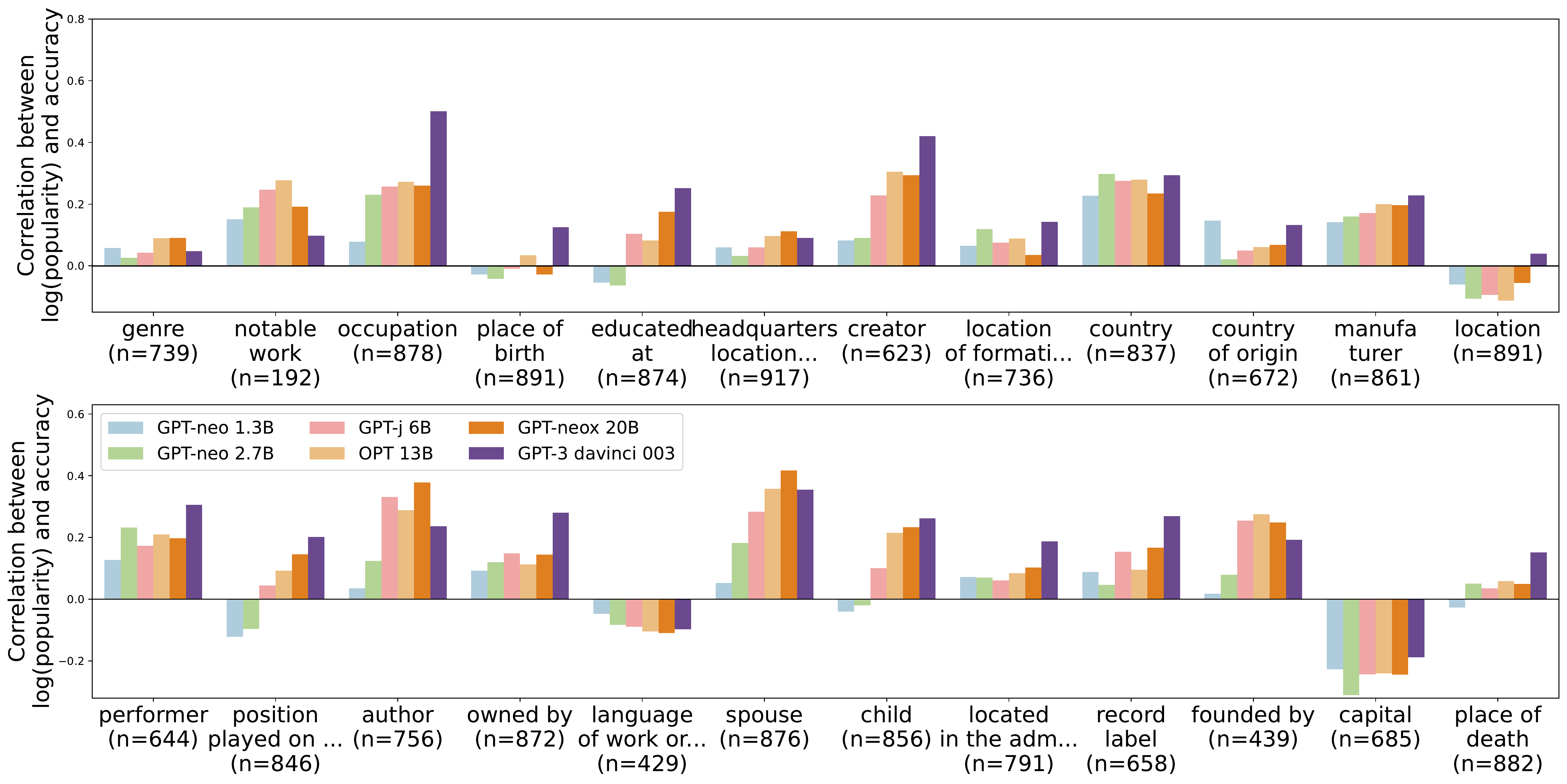}
    \caption{Correlations on EntityQuestions for all relationship types and models.}
    \label{fig:appendix_EQ_accuracy_breakdown}
\end{figure*}

\paragraph{{Negative correlations of capital on EntityQuestions.}}
{As shown in Figure~\ref{fig:appendix_EQ_accuracy_breakdown}, the capital relationship types on in EntityQuestions, while on \ours, this relationship shows relatively high correlations. We found that in EntityQuestions, this capital relationship type has many low-popularity questions whose answers are included in subject entity names (e.g., subject="canton of Marseille-Belsunce", object="Marseille").} 
{This causes performance to have a U-shaped relationship with popularity for the capital relationship type, so if most of the questions sampled come from the top half of popularity, the linear correlation will be positive, and vice versa.}

\subsection{Retrieval-augmented LM results} \label{app:retr_results}

\paragraph{Overall performance of retrieval-augmented LMs. }
Figure~\ref{fig:appendix_acc_by_model} shows the overall performance of 40 LMs and retrieval-augmented LMs on \ours. 
Retrieval-augmentation largely improves performance across different LMs, and much smaller models (GPT-Neo 1.3B) can perform on per with GPT-3. 
{Figure~\ref{fig:appendix_EQ_acc_by_model} shows the results on EntityQuestions. } Due to computational and time constraints, we were only able to run vanilla and Contriever results for most models.

\paragraph{Adaptive Retrieval for EntityQuestions.} Figure~\ref{fig:appendix_EQ_thresh_by_model} shows the proportion of questions above the retrieval threshold for various models using Adaptive Retrieval on EntityQuestions. Because EntityQuestions has a large quantity of low-popularity questions, models (especially smaller ones) must rely heavily on retrieval.

\paragraph{Full results on all relationship types.}
Figure~\ref{fig:appendix_pop_linecharts} shows the full results on \ours\ of the retrieval-augmented LMs and unassisted LMs on 16 relationship types using three different LMs as backbones. Figure~\ref{fig:appendix_EQ_linecharts} shows these results for GPT-3 \texttt{davinci-003} on EntityQuestions.

\begin{figure*}
    \centering
    \begin{overpic}[width=0.95\textwidth]{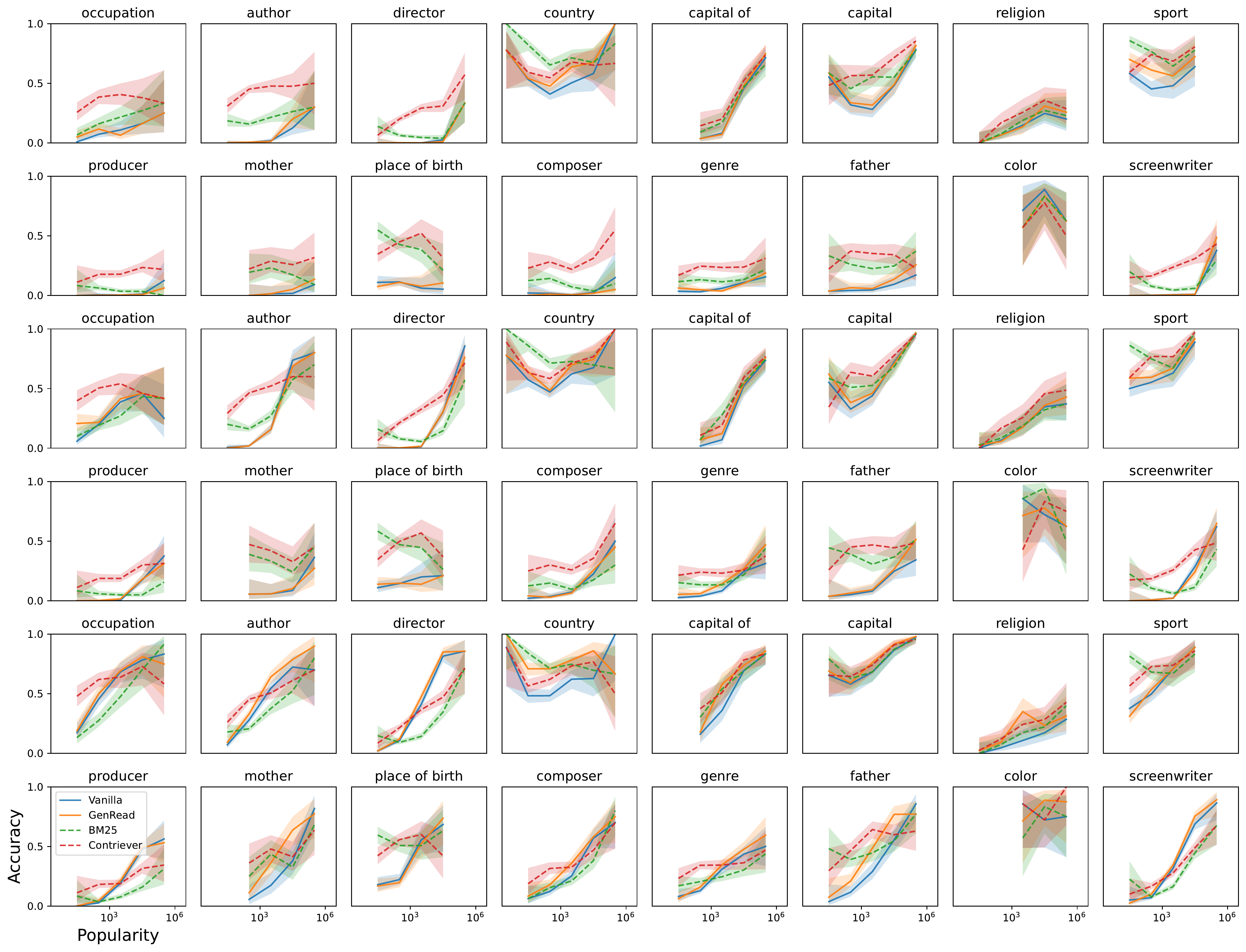}
        \put(100.5, 70){\rotatebox{-90}{\small{GPT-neo 1.3B}}}
        \put(100.5, 44){\rotatebox{-90}{\small{OPT 13B}}}
        \put(100.5, 22){\rotatebox{-90}{\small{GPT-3 DaVinci 003}}}
    \end{overpic}
    \caption{Accuracy for 3 models on \ours\ versus popularity as shown in Figure~\ref{fig:retrieval_how_helps} broken down by relationship type. Popularity bins with less than 5 samples are excluded to avoid cluttering the figures with noisy results that have wide error bars.}
    \label{fig:appendix_pop_linecharts}
\end{figure*}

\begin{figure*}
    \centering
    \includegraphics[width=0.95\textwidth]{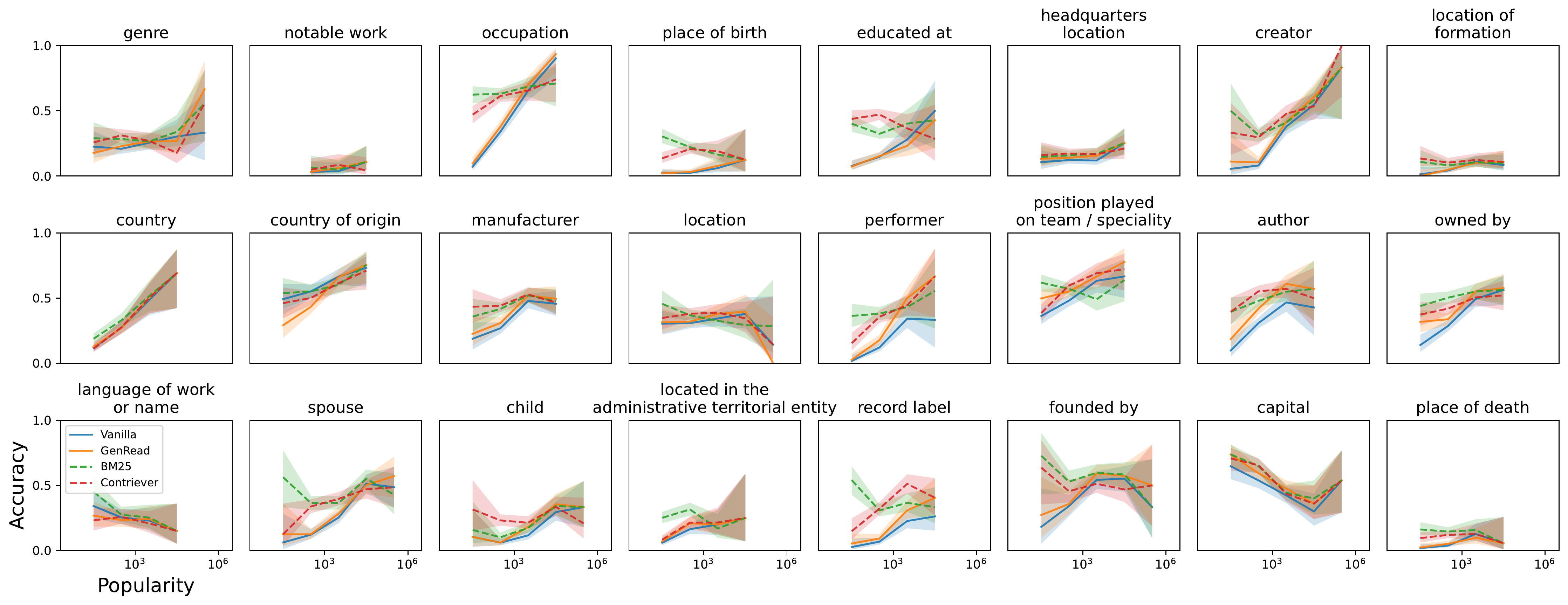}
    \caption{Accuracy versus popularity for GPT-3 \texttt{davinci-003} on EntityQuestions broken down by relationship type. Popularity bins with less than 5 samples are excluded.}
    \label{fig:appendix_EQ_linecharts}
\end{figure*}

\subsection{Qualitative Results}
\label{sec:qualitatitve_results}

Table~\ref{tab:group_b} shows several examples on \ours, where GPT-3 \texttt{davinci-003} answers correctly while the Contriever-augmented version fails to answer. Along with the low recall@1 of 0.14 for this group, Table~\ref{tab:group_b} suggests that the most common reason retrieval can be harmful is that it retrieves a document about a mistaken entity, such as a person with the same name as the subject, or an entity that simply is not relevant to the question (as in the case of ``Noel Black'').

Table~\ref{tab:group_c} shows several examples on \ours, where GPT-3 \texttt{davinci-003} answers correctly only when augmented with Contriever. The recall@1 for this case is 0.88, which is significantly higher than the overall recall. Note that in the second example, the retrieval caused the LM to answer correctly, but only by coincidence: the subject entity ``Pierre'' actually refers to the city in South Dakota, not the Basketball player. Otherwise, retrieval appears to be helpful because it provides the relevant information directly.

\begin{table*}[]
    \centering
    \small
    \begin{tabular}{p{3cm}p{2cm}p{2cm}p{7cm}}
    \toprule
         Question (answer) & LM  & Retrieval-LM & Retrieval  \\\midrule
         Who was the director of Black? (Sanjay Leela Bhansali) &
The director of Black is Sanjay Leela Bhansali. \color{green}{\checkmark} &
 Noel Black \color{red}{\xmark} & 
Noel Black (June 30, 1937 – July 5, 2014) was an American film and television director, screenwriter, and producer. Black was born in Chicago, Illinois. He won awards at the 1966 ... and subsequently concentrated on directing for television, occasionally directing films such as "Private School".\\
\\
         Who was the producer of The Faculty? (Robert Rodriguez) 
 & The Faculty was produced by Elizabeth Avellan and Robert Rodriguez. \color{green}{\checkmark} &  The producer of The Faculty was Elizabeth Avellan. \textcolor{red}{\xmark} &  The Faculty is a 1998 American science fiction teen horror film written by Kevin Williamson, directed by \textcolor{blue}{\bf \underline {Robert Rodriguez}}, and starring Elijah Wood,... The film was released on Christmas Day, 1998, by Miramax Films through Dimension Films. ... The film has developed a cult following in the years since its release.  \\ 
 \\
 What is Michael Shelley's occupation?
 (singer-songwriter) & Michael Shelley is a singer-songwriter and musician. \color{green}{\checkmark} & 
 Michael Shelley is an American applied mathematician. \color{red}{\xmark} &
Michael Shelley (mathematician) Michael J. Shelley (born August 17, 1959) is an American applied mathematician who works on the modeling and simulation of complex systems arising in physics and biology. This has included free-boundary problems in fluids and materials science... He is also the co-founder and co-director of the Courant Institute's Applied Mathematics Lab. \\
\\
In what city was Zijah Sokolović born? (Sarajevo) & 
Zijah Sokolović was born in Sarajevo \color{green}{\checkmark} & Zijah Sokolović was born in Orahovac, Kingdom \color{red}{\xmark} & Ali Sokol ... (born 8 May 1921 in Orahovac, Kingdom of Serbs, Croats and Slovenes, died 23 September 1974) was a Yugoslav pulmonologist . Ali Sokol was born into an agricultural family. He was the third of four children of father and mother Hatixhes Solomon. It is unknown the exact date of birth but the most reliable date is May 8 year in 1921.
\\ 
\bottomrule
    \end{tabular}
    \caption{Qualitative examples of the questions where only retrieval-augmented LMs \textit{fail} to answer correctly. The blue underlined text indicates the sub-strings matching the gold answers in the retrieved context. }
    \label{tab:group_b}
\end{table*}

\begin{table*}[]
    \centering
    \small
    \begin{tabular}{p{3cm}p{2cm}p{2cm}p{7cm}}
    \toprule
         Question (answer) & LM  & Retrieval-LM & Retrieval  \\\midrule
         What genre is Unknown? (fantasy) &
Unknown is not a specific genre of music. It could refer to \color{red}{\xmark} &
 Unknown is a pulp fantasy fiction magazine. \color{green}{\checkmark} &
Unknown (magazine) Unknown (also known as Unknown Worlds) was an American pulp \textcolor{blue}{\bf \underline {fantasy}} fiction magazine, published from 1939 to 1943 by Street \& Smith, and edited by John W. Campbell. ... The leading fantasy magazine in the 1930s was "Weird Tales", which focused on shock and horror.\\
\\
         In what country is Pierre? (United States) &
Pierre is in France. \color{red}{\xmark} & 
 Pierre is from the United States. \color{green}{\checkmark} & 
Jonas Farley Pierre (born May 30, 1980) is an American professional basketball player. A center, he has played with three teams in the National Basketball League of Canada (NBL). Pierre last played with the Saint John Mill Rats in Canada. ... He then joined Lincoln University in Missouri for his third year. In the summer of 2003, he signed a letter of intent to play with Nicholls State in the NCAA Division I.  \\ 
\\
Who was the producer of The Cocoanuts? (Walter Wanger)
 & The Cocoanuts was produced by Florenz Ziegfeld. \color{red}{\xmark} &
 The Cocoanuts was produced for Paramount Pictures by Walter Wanger, who \color{green}{\checkmark} &
The Cocoanuts is a 1929 musical comedy film starring the Marx Brothers. Produced for Paramount Pictures by \textcolor{blue}{\bf \underline {Walter Wanger}}, who is not credited, the film stars the four Marx Brothers, Oscar Shaw, Mary Eaton, and Margaret Dumont. It was the first sound film to credit more than one director (Robert Florey and Joseph Santley), and was adapted to the screen by Morrie Ryskind from the George S. Kaufman Broadway musical play. ... \\
\\
Who was the director of The White Suit?
(Lazar Ristovski) &
The White Suit was directed by Sachin Kundalkar. \color{red}{\xmark} &
 Lazar Ristovski \color{green}{\checkmark}
& In 1999 "The White Suit" an auteur film by Ristovski (director, writer, lead actor, and producer) was at the Cannes Film Festival in the Critics Week program. "The White Suit" was the Serbian entry for the 1999 Academy Awards. \textcolor{blue}{\bf \underline {Lazar Ristovski}} is the sole owner of Zillion Film Company In 2006, he made a small appearance in the James Bond film "Casino Royale". He played Caruso in the 2004 movie "King of Thieves". He starred as Đorđe in the award-winning 2009 film "St. George Shoots the Dragon".
\\ 
\bottomrule
    \end{tabular}
    \caption{Qualitative examples of the questions where only retrieval-augmented LMs \textit{successfully} answer correctly.
    The blue underlined text indicates the sub-strings matching the gold answers in the retrieved context. }
    \label{tab:group_c}
\end{table*}

\end{document}